\documentclass[conference]{IEEEtran}
\IEEEoverridecommandlockouts
\usepackage{amsmath,amssymb,amsfonts}
\usepackage{graphicx}
\usepackage{textcomp}
\usepackage{xcolor}
\usepackage{booktabs}
\usepackage{hyperref}
\usepackage{comment}
\usepackage[numbers]{natbib} 

\title{PARWiS: Winner determination under shoestring budgets using active pairwise comparisons}

\author{
    \IEEEauthorblockN{Shailendra Bhandari}
    \IEEEauthorblockA{Department of Computer Science\\
    OsloMet-- Oslo Metropolitan University
    Oslo, Norway\\
    Email: \url{shailendra.bhandari@oslomet.no}}
}

\begin{document}

\maketitle

\begin{abstract}
Determining a winner among a set of items using active pairwise comparisons under a limited budget is a challenging problem in preference-based learning. The goal of this study is to implement and evaluate the PARWiS algorithm, which shows spectral ranking and disruptive pair selection to identify the best item under shoestring budgets. This work have extended the PARWiS with a contextual variant (Contextual PARWiS) and a reinforcement learning-based variant (RL PARWiS), comparing them against baselines, including Double Thompson Sampling and a random selection strategy. This evaluation spans synthetic and real-world datasets (Jester and MovieLens), using budgets of 40, 60, and 80 comparisons for 20 items. The performance is measured through recovery fraction, true rank of reported winner, reported rank of true winner, and cumulative regret, alongside the separation metric \(\Delta_{1,2}\). Results show that PARWiS and RL PARWiS outperform baselines across all datasets, particularly in the Jester dataset with a higher \(\Delta_{1,2}\), while performance gaps narrow in the more challenging MovieLens dataset with a smaller \(\Delta_{1,2}\). Contextual PARWiS shows comparable performance to PARWiS, indicating that contextual features may require further tuning to provide significant benefits.
\end{abstract}


\section{Introduction}
Preference-based learning through pairwise comparisons is a powerful approach for ranking and winner determination in applications such as recommender systems, social choice, and information retrieval \cite{10.1007/978-3-319-91908-9_4,10.1007/s10994-019-05867-2}. However, in real-world scenarios, the number of comparisons is often constrained by a limited budget, known as a shoestring budget. Sheth and Rajkumar \cite{sheth2021parwis} introduced PARWiS (Pairwise Active Recovery of Winner under a Shoestring budget), an algorithm designed to efficiently determine the winner under such constraints using spectral ranking and a disruptive pair selection strategy.

In this work, PARWiS has been implemented and extended it with two variants: Contextual PARWiS, inspired by Bengs et al. \cite{xu2024linearcontextualbanditsinterference, datsai2022fastonlineinferencenonlinear}, to incorporate item features when available, and RL PARWiS, a reinforcement learning-based approach that uses Q-learning to optimize pair selection. these algorithms has been compared against two baselines: Double Thompson Sampling (Double TS) \cite{NIPS2016_9de6d14f} and a random pair selection strategy. The evaluation uses synthetic data generated via the Bradley-Terry (BT) model, as well as real-world datasets: Jester \cite{goldberg2001eigentaste} and MovieLens 20M \cite{harper2015movielens}. The performance has been ascessed across three shoestring budgets (40, 60, and 80 comparisons for 20 items) using metrics from \cite{sheth2021parwis}: recovery fraction, true rank of reported winner, reported rank of true winner, and cumulative regret. Additionally, the separation metric \(\Delta_{1,2}\) is computed to analyze problem difficulty and perform statistical tests to assess the significance of performance differences.

The results demonstrate that PARWiS and RL PARWiS consistently outperform the baselines in terms of recovery fraction and cumulative regret, particularly on the Jester dataset, where a higher \(\Delta_{1,2}\) indicates an easier problem. On the more challenging MovieLens dataset with a smaller \(\Delta_{1,2}\), PARWiS and RL PARWiS still perform better but with a narrower margin. Contextual PARWiS performs similarly to PARWiS, suggesting that the contextual features used (available only in synthetic data) may require further optimization. Statistical tests confirm that PARWiS and RL PARWiS’s improvements over baselines are significant, while error analysis reveals that RL PARWiS fails closer to the true winner compared to other agents.

\section{Related work}
The problem of winner determination and ranking through pairwise comparisons has been extensively studied in the dueling bandits framework, which bridges multi-armed bandits and preference-based learning. This framework is particularly relevant in applications such as recommender systems, social choice, and information retrieval, where direct numerical feedback is often unavailable, and preferences must be inferred through comparisons. Below, we review key developments in dueling bandits, active learning for ranking, contextual extensions, and the use of real-world datasets for evaluation.

\subsection{Dueling bandits and regret minimization}
The dueling bandits framework was introduced to model scenarios where feedback is relative rather than absolute, typically in the form of pairwise comparisons. Early work by Yue et al. \cite{YUE20121538} formalized the \( k \)-armed dueling bandits problem, focusing on regret minimization. They proposed algorithms like the Interleaved Filter, which balances exploration and exploitation by maintaining a set of potential winners and refining it through comparisons. Their regret bounds scale with the number of arms and the comparison budget, but the approach assumes a Condorcet winner exists, which may not hold in all settings.

\begin{figure*}[t]
    \centering
    \includegraphics[width=0.22\textwidth]{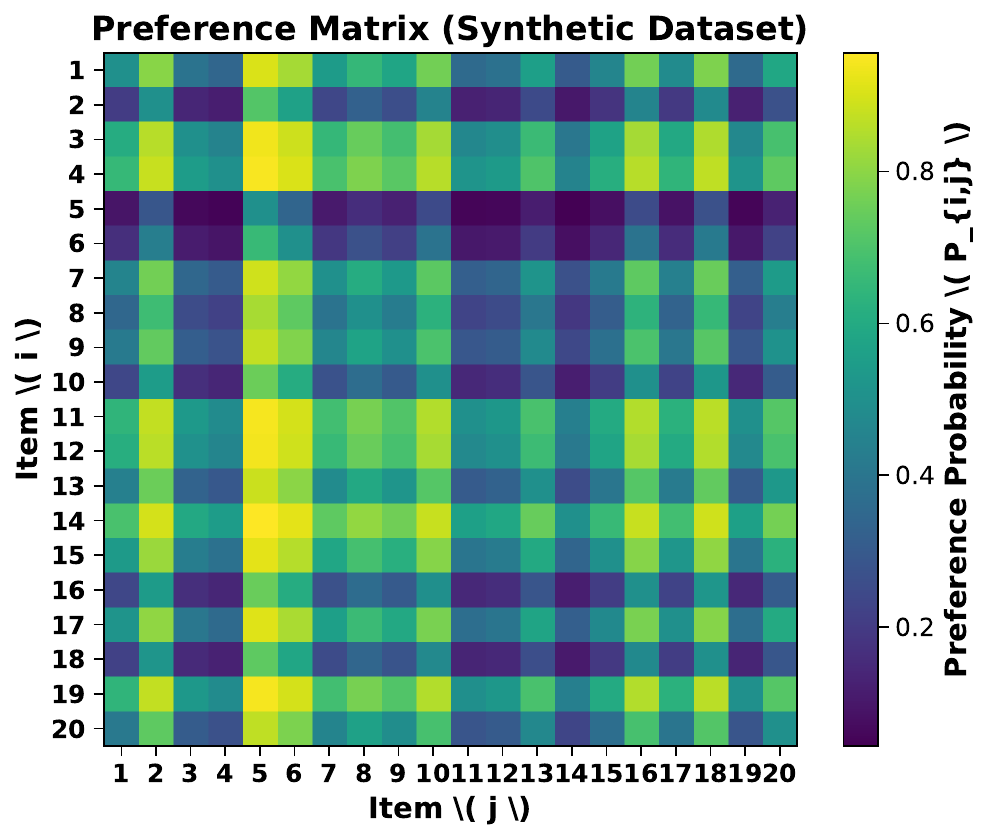}
    \includegraphics[width=0.24\textwidth]{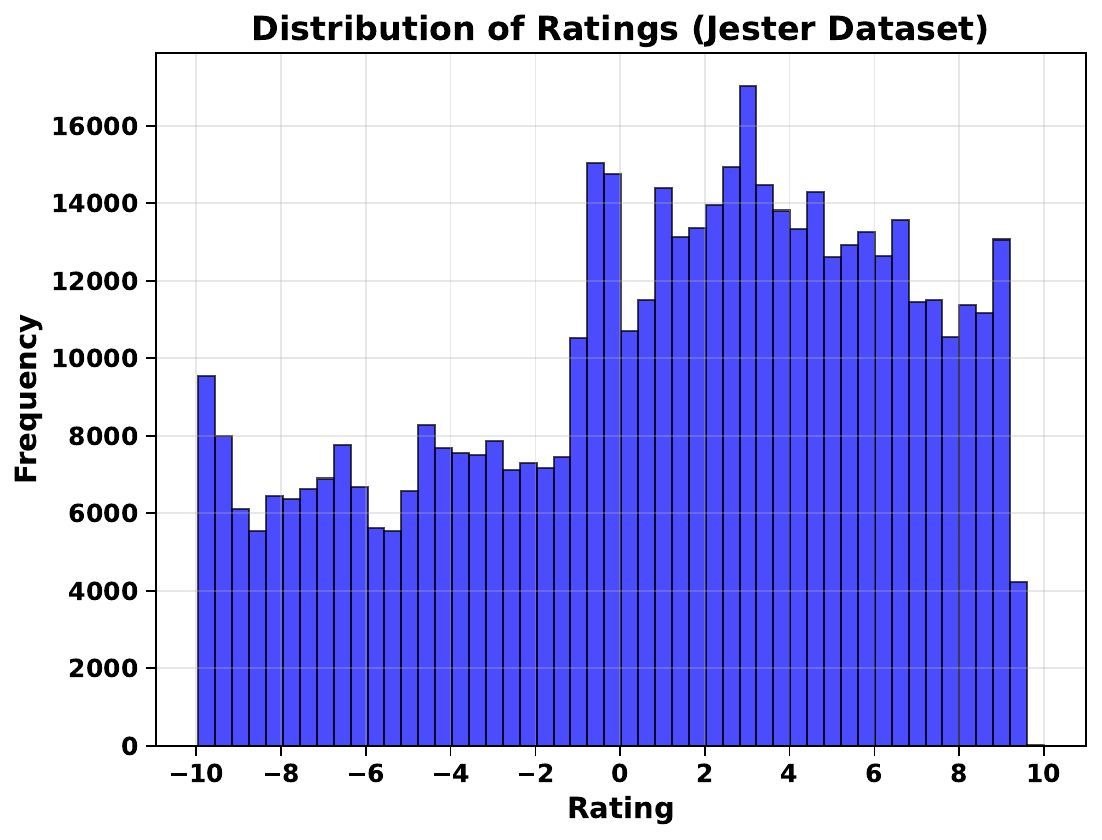}
    \includegraphics[width=0.24\textwidth]{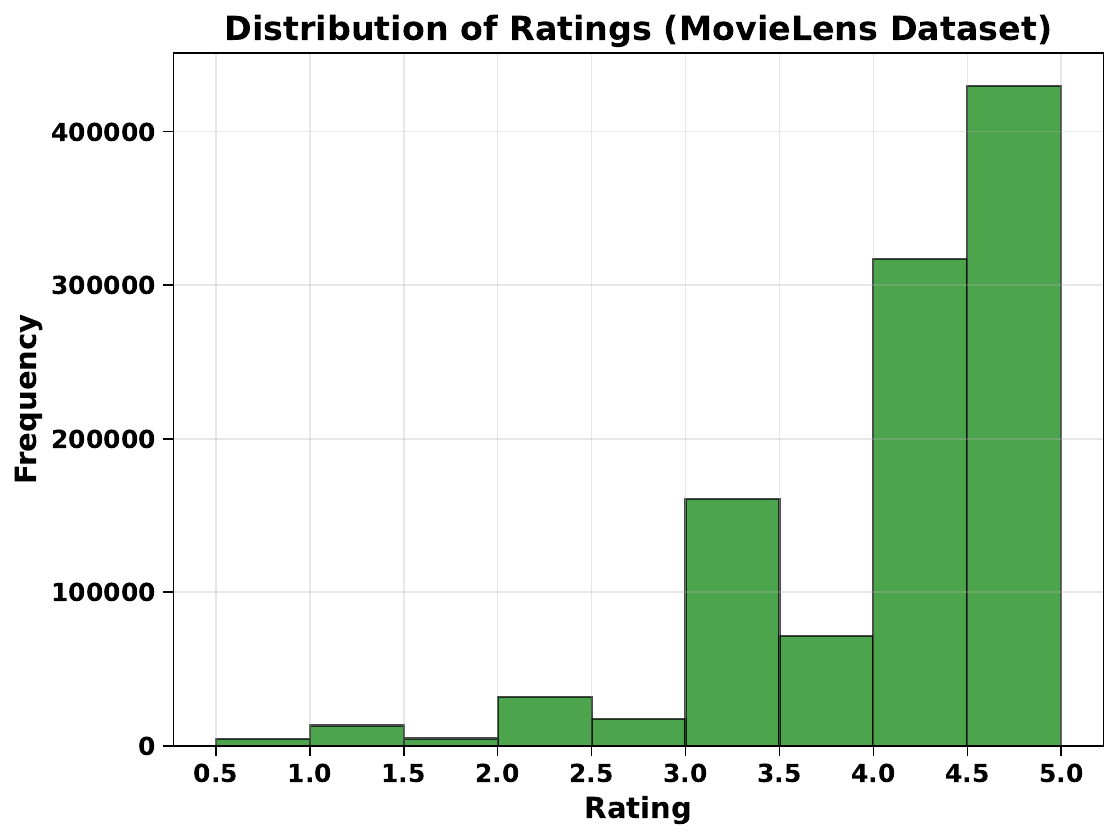}
    \includegraphics[width=0.27\textwidth]{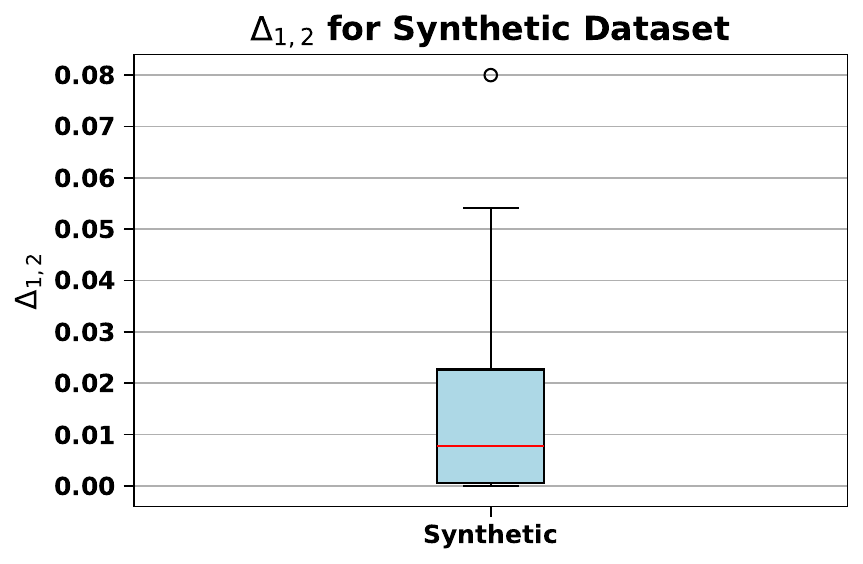}
    \caption{Dataset Visualizations: (a) Heatmap of the preference matrix for the Synthetic dataset (\( k=20 \)), showing pairwise probabilities \( P_{i,j} \). (b) Histogram of ratings for the 20 selected jokes in Jester, ranging from -10 to 10. (c) Histogram of ratings for the 20 selected movies in MovieLens, ranging from 0.5 to 5. (d) Boxplot of \(\Delta_{1,2}\) for the Synthetic dataset across 30 runs, showing the distribution of problem difficulty.}
    \label{fig:dataset_visualizations}
\end{figure*}
Subsequent works extended this framework to handle more complex preference structures. Zoghi et al. \cite{zoghi2014relative} introduced the Relative Upper Confidence Bound (RUCB) algorithm, which adapts the Upper Confidence Bound (UCB) strategy to dueling bandits. RUCB maintains confidence intervals over pairwise preferences and selects pairs that maximize the potential for reducing uncertainty about the winner. While RUCB achieves sublinear regret, it requires a large number of comparisons to converge, making it less suitable for shoestring budgets.

Sparse Borda \cite{Jamieson2015SparseDB} further refines this approach by incorporating sparsity conditions in the pairwise comparison matrix to eliminate suboptimal items early. This algorithm targets the Borda winner (the item with the highest average preference score), which aligns with the Bradley-Terry-Luce (BTL) winner in our setting. However, its performance under shoestring budgets is limited, as shown by Sheth and Rajkumar \cite{sheth2021parwis}, due to insufficient comparisons to accurately estimate the sparse structure.

\subsection{Active learning for ranking and winner determination}
Active learning approaches aim to minimize the number of comparisons needed to rank items or identify the top-\( k \) items. Jamieson and Nowak \cite{jamieson2011activerankingusingpairwise} proposed an active ranking algorithm that models items as embeddings in a \( d \)-dimensional Euclidean space, using geometric properties to select informative pairs. While effective for noiseless comparisons, their method struggles with noisy models like BTL, which is central to our study.

Mohajer et al. \cite{pmlr-v70-mohajer17a} introduced SELECT, a single-elimination tournament algorithm for top-\( k \) rank aggregation. Select pairs of items randomly at each layer, promoting winners to the next layer until the top-\( k \) items are identified. The algorithm is extended to handle shoestring budgets by repeating comparisons within layers, but its random pairing strategy limits its efficiency, as noted in \cite{sheth2021parwis}. Heckel et al. \cite{heckel2019active} proposed Active Ranking (AR), a generic framework for full ranking and winner determination. AR maintains confidence scores for items, partitioning them into bins to focus comparisons on uncertain pairs. While AR performs well with parametric assumptions (e.g., BTL), it requires more comparisons than shoestring budgets typically allow, making it less competitive in our setting.

Sheth and Rajkumar \cite{sheth2021parwis} introduced PARWiS (Pairwise Active Recovery of Winner under a Shoestring budget), specifically targeting winner determination under constrained budgets. PARWiS leverages spectral ranking \cite{negahban2016rank} to estimate BTL scores and selects the most disruptive pairs to update the ranking iteratively. The algorithm consists of an initialization phase (\( k-1 \) comparisons to build an initial ranking) and an update phase that uses a disruption measure to choose pairs that maximize the change in the ranking. PARWiS outperforms baselines like SELECT and AR in shoestring settings, as demonstrated on synthetic and real-world datasets. Our work builds on PARWiS by implementing and extending it with a contextual variant.

\subsection{Contextual dueling bandits}
Contextual dueling bandits incorporate item features to improve pair selection, particularly in settings with abundant comparisons. Bengs et al. \cite{xu2024linearcontextualbanditsinterference, datsai2022fastonlineinferencenonlinear} provide a survey of contextual bandits, highlighting algorithms that use linear or nonlinear models to predict comparison outcomes based on features. For example, Contextual Dueling Bandits with Linear Payoffs (CBLP) \cite{yue2009interactively} models preferences as a linear function of item features, selecting pairs that maximize expected information gain. While effective, these methods often assume access to rich feature sets, which may not be available in real-world datasets like Jester or MovieLens.

More recent works explore nonlinear models for greater flexibility. Saha and Gopalan \cite{saha2018contextual} propose a kernel-based approach to model complex preference relationships, achieving tighter regret bounds in high-dimensional settings. However, these methods require significant computational resources and a large number of comparisons, making them impractical for shoestring budgets. In our work, we extend PARWiS with Contextual PARWiS, which uses logistic regression to predict comparison outcomes based on features. However, since real-world datasets lack features, Contextual PARWiS falls back to non-contextual behavior in those cases.

\subsection{Spectral ranking and preference aggregation}
Spectral ranking methods estimate item scores from pairwise comparisons, forming the backbone of algorithms like PARWiS. Negahban et al. \cite{negahban2016rank} introduced Rank Centrality, which constructs an augmented Markov chain from comparison outcomes and computes its stationary distribution to approximate BTL scores. Rank Centrality requires only \( \mathcal{O}(n \log n) \) comparisons to achieve reasonable error bounds, making it suitable for active learning settings. Spectral MLE \cite{chen2015spectral} refines these estimates using coordinate-wise maximum likelihood estimation, improving accuracy for top-\( k \) ranking. However, these methods are typically passive, assuming a fixed set of comparisons, whereas PARWiS actively selects pairs to optimize the ranking.

Agarwal et al. \cite{agarwal2018accelerated} proposed an accelerated version of spectral ranking, reducing computational complexity by orders of magnitude. While promising, its application in active settings remains underexplored, and we leave its integration into PARWiS for future work. Other preference aggregation methods, such as Copeland Aggregation \cite{saari1996copeland} used in Multisort \cite{maystre2017just}, combine rankings from multiple sources (e.g., noisy quicksorts) to determine a winner. These methods are less effective under shoestring budgets, as they require multiple full rankings, which exceed the comparison limit.

\subsection{Real-world datasets for preference learning}
Real-world datasets provide practical benchmarks for evaluating preference-based learning algorithms. The Jester dataset \cite{goldberg2001eigentaste}, collected between 1999 and 2003, contains 4.1 million ratings of 100 jokes from 73,421 users, with ratings ranging from -10 to +10. It has been widely used in recommender systems research due to its dense rating matrix, particularly for a subset of "gauge" jokes rated by most users. MovieLens 20M \cite{harper2015movielens}, with 20 million ratings for 27,000 movies from 138,000 users, is another standard benchmark, offering a sparse rating matrix that challenges algorithms to handle missing data. Both datasets have been used in \cite{sheth2021parwis} to evaluate PARWiS, converting ratings to pairwise probabilities using a logistic function on rating differences, a method we adopt in our experiments.

Other datasets, such as Sushi \cite{10.1145/956750.956823}, provide preference rankings directly, which can be converted to pairwise comparisons. Crowd-sourcing approaches, like Crowd-BT \cite{10.1145/2433396.2433420}, extend the BTL model to handle malicious user behavior, modeling voter reliability alongside item scores. These datasets and methods highlight the diversity of preference data in real-world applications, motivating our evaluation on both synthetic and real-world datasets.

\subsection{Challenges in shoestring budgets}
Shoestring budgets pose unique challenges for dueling bandits and active ranking. Algorithms like RUCB and Sparse Borda require many comparisons to achieve low regret, often infeasible under constraints like \( B = 2k \). Active learning methods such as SELECT and AR struggle with random or suboptimal pair selection, leading to poor recovery rates. PARWiS addresses these challenges by focusing on disruptive pairs that maximize ranking updates, a strategy we replicate and extend in our work. However, the impact of problem difficulty (measured by \(\Delta_{1,2}\)) remains a critical factor, as smaller separations between top items make winner determination harder, a phenomenon we explore in our experiments.

In summary, this work builds on the foundation laid by PARWiS, incorporating contextual and reinforcement learning extensions and evaluating performance across diverse datasets. By addressing the limitations of existing methods under shoestring budgets, this work will contribute to the understanding of active winner determination in preference-based learning.

\section{Methodology}
\subsection{Problem setting}
Consider a set of \( k \) items, where pairwise comparisons are performed under the Bradley-Terry-Luce (BTL) model. In the BTL model, each item \( i \) has a score \( w_i \), and the probability that item \( i \) beats item \( j \) is given by:
\[
P_{i,j} = \frac{w_i}{w_i + w_j}.
\]
Given a budget \( B \), the goal is to identify the item with the highest score (the winner) by performing at most \( B \) comparisons. We evaluate performance using shoestring budgets \( B = 2k, 3k, 4k \), following \cite{sheth2021parwis}.

\subsection{Algorithms}
Here I have implemented five algorithms:
\begin{itemize}
    \item \textbf{Double Thompson Sampling (Double TS)} \cite{NIPS2016_9de6d14f}: Uses two Thompson Sampling steps to select pairs, maintaining Beta priors over pairwise preferences.
    \item \textbf{Random}: Selects pairs uniformly at random, serving as a baseline.
    \item \textbf{PARWiS} \cite{sheth2021parwis}: Employs an initialization phase with \( k-1 \) comparisons to build a spectral ranking, followed by a phase that selects the most disruptive pairs to update the ranking.
    \item \textbf{Contextual PARWiS}: The extension of PARWiS, which incorporates item features (when available) to predict comparison outcomes using logistic regression, inspired by \cite{xu2024linearcontextualbanditsinterference, datsai2022fastonlineinferencenonlinear}.
    \item \textbf{RL PARWiS}: A reinforcement learning-based extension of PARWiS, using Q-learning to learn a pair selection policy. The state includes the current ranking and comparison counts, the action is the choice of a pair to compare, and the reward combines regret reduction per step with a final reward for recovering the true winner.
\end{itemize}

\subsection{Datasets}
The algorithms has been evaluated on three datasets, providing detailed characteristics and visualizations to highlight their properties.

\begin{itemize}
    \item \textbf{Synthetic}: Generated using the BTL model with \( k=20 \) items and \( d=5 \) feature dimensions. Item scores \( w_i \) are sampled from a normal distribution, and pairwise probabilities are computed as \( P_{i,j} = w_i / (w_i + w_j) \). Features are random normal vectors, used only by Contextual PARWiS. Figure~\ref{fig:dataset_visualizations}(a) shows a heatmap of the preference matrix \( P_{i,j} \), illustrating the variability in pairwise preferences. Figure~\ref{fig:dataset_visualizations}(d) shows a boxplot of \(\Delta_{1,2}\) across 30 runs, confirming the reported variability (\(\Delta_{1,2} = 0.0152 \pm 0.0190\)) and indicating moderate difficulty with some runs being easier or harder.

    \item \textbf{Jester}: Jester Dataset 1 (subset 1) \cite{goldberg2001eigentaste}, containing 4.1 million ratings for 100 jokes from 73,421 users, collected between 1999 and 2003. Ratings range from -10 to +10, with 99 indicating missing ratings. I have selected a random subset of 20 jokes (using a fixed seed for reproducibility) to align with the difficulty reported in \cite{sheth2021parwis} (\(\Delta_{1,2} = 0.0043\)). However, the selection yields \(\Delta_{1,2} = 0.0946 \pm 0.0000\), suggesting a relatively easier problem, possibly due to the distribution of ratings. Figure~\ref{fig:dataset_visualizations}(b) shows a histogram of ratings for the selected jokes, revealing a slightly right-skewed distribution centered around 0, with most ratings between -5 and 5. This dense rating matrix facilitates accurate estimation of pairwise probabilities.

    \item \textbf{MovieLens}: MovieLens 20M \cite{harper2015movielens}, with 20 million ratings for 27,000 movies from 138,000 users, collected between 1995 and 2015. Ratings range from 0.5 to 5 (in increments of 0.5). We select the top 20 movies by number of ratings, resulting in a sparse rating matrix with \(\Delta_{1,2} = 0.0008 \pm 0.0000\), indicating a challenging problem where the top two movies are nearly indistinguishable. Figure~\ref{fig:dataset_visualizations}(c) shows a histogram of ratings for the selected movies, displaying a left-skewed distribution with most ratings between 3 and 5, reflecting a tendency for users to rate popular movies positively.
\end{itemize}

For real-world datasets, ratings are converted to pairwise probabilities using a logistic function on the difference of average ratings, following \cite{sheth2021parwis}. The absence of item features in Jester and MovieLens means Contextual PARWiS falls back to non-contextual behavior for these datasets.

\begin{figure*}[t]
    \centering
    \includegraphics[width=0.24\textwidth]{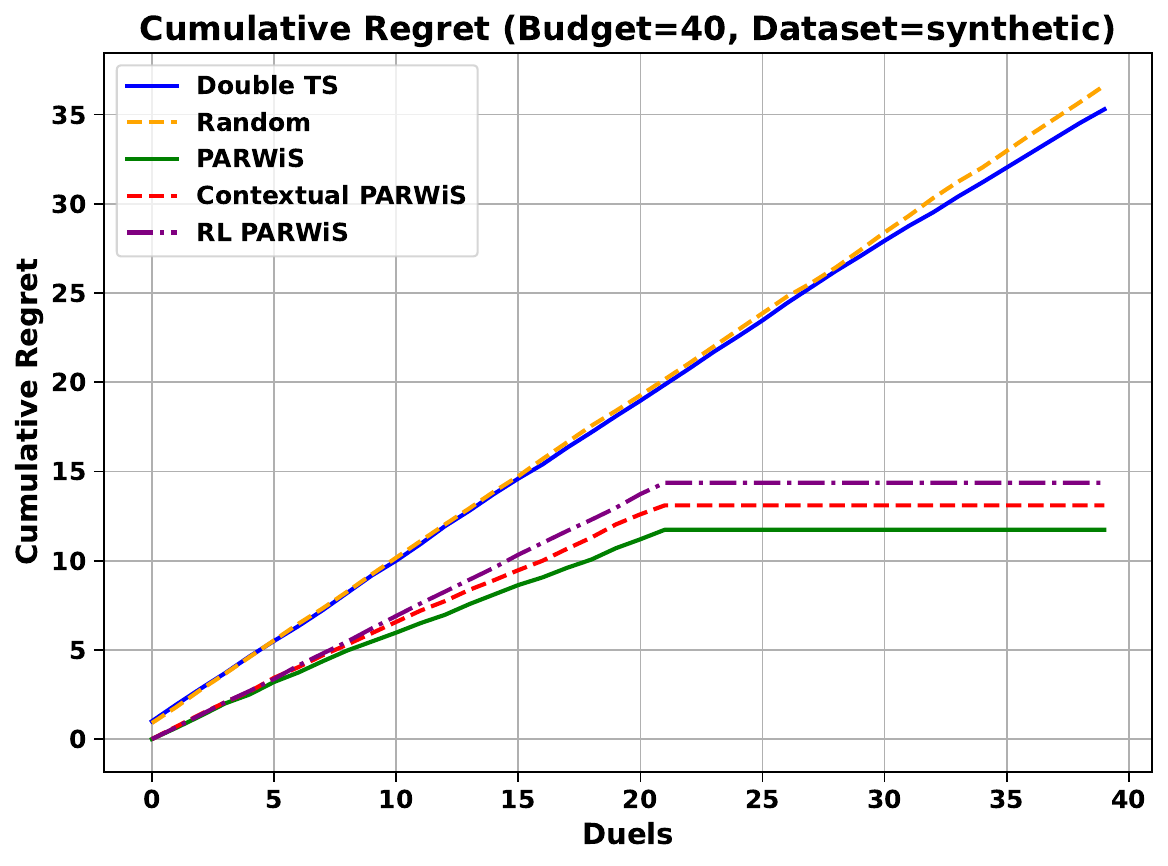}
    \includegraphics[width=0.24\textwidth]{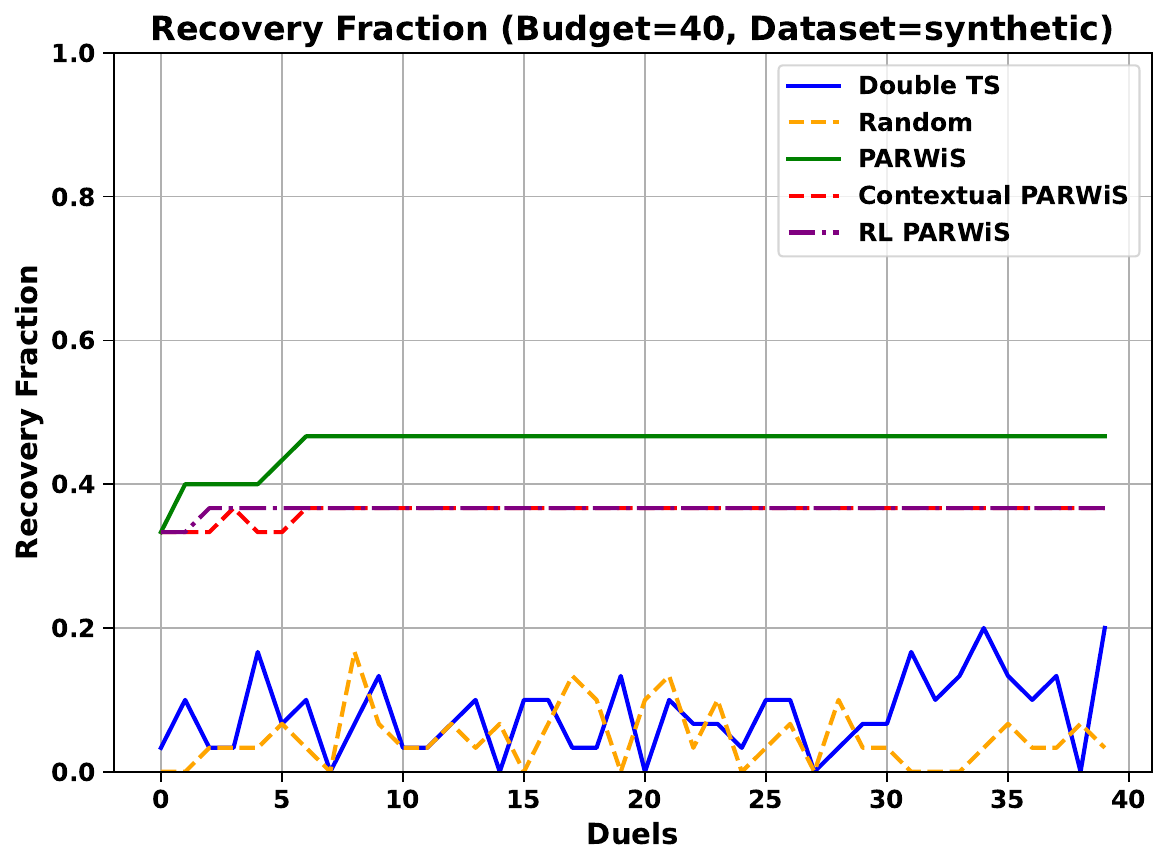}
    \includegraphics[width=0.25\textwidth]{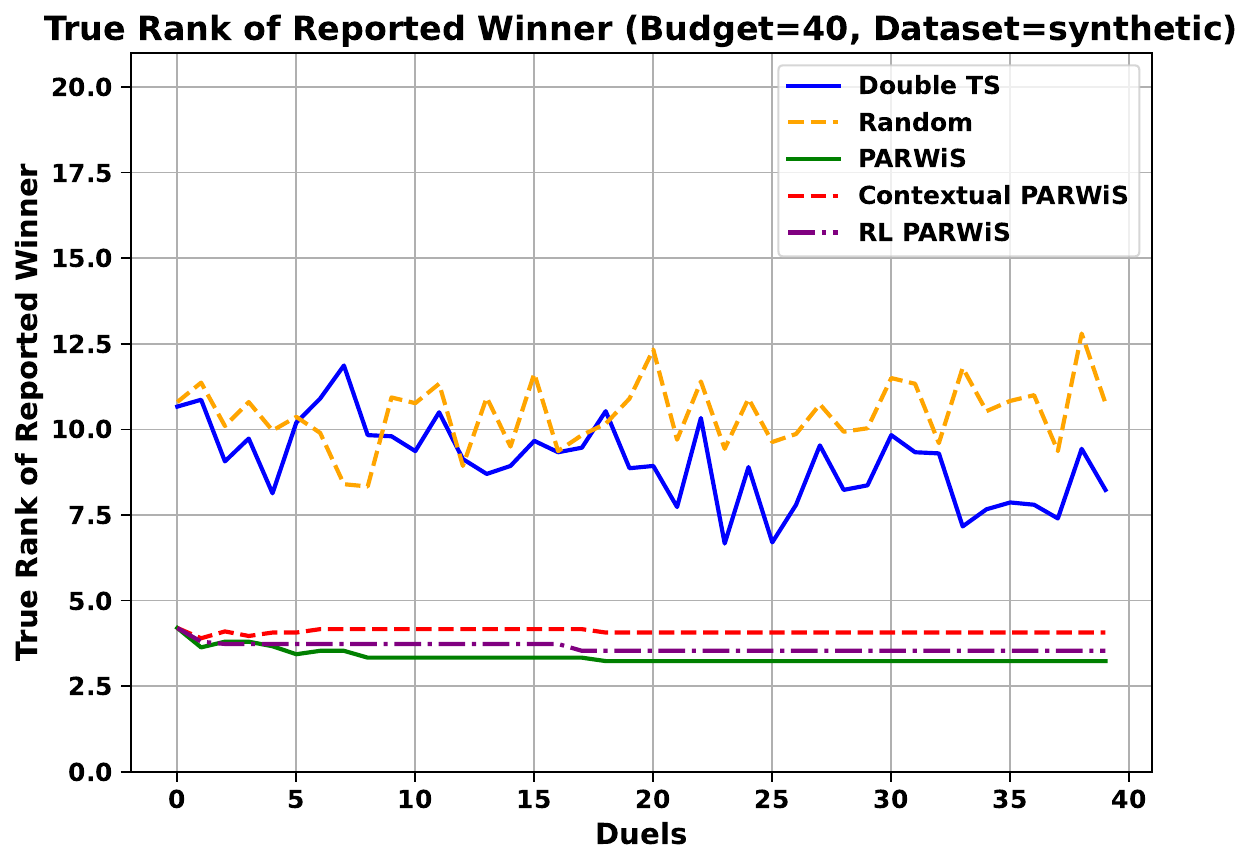}
    \includegraphics[width=0.25\textwidth]{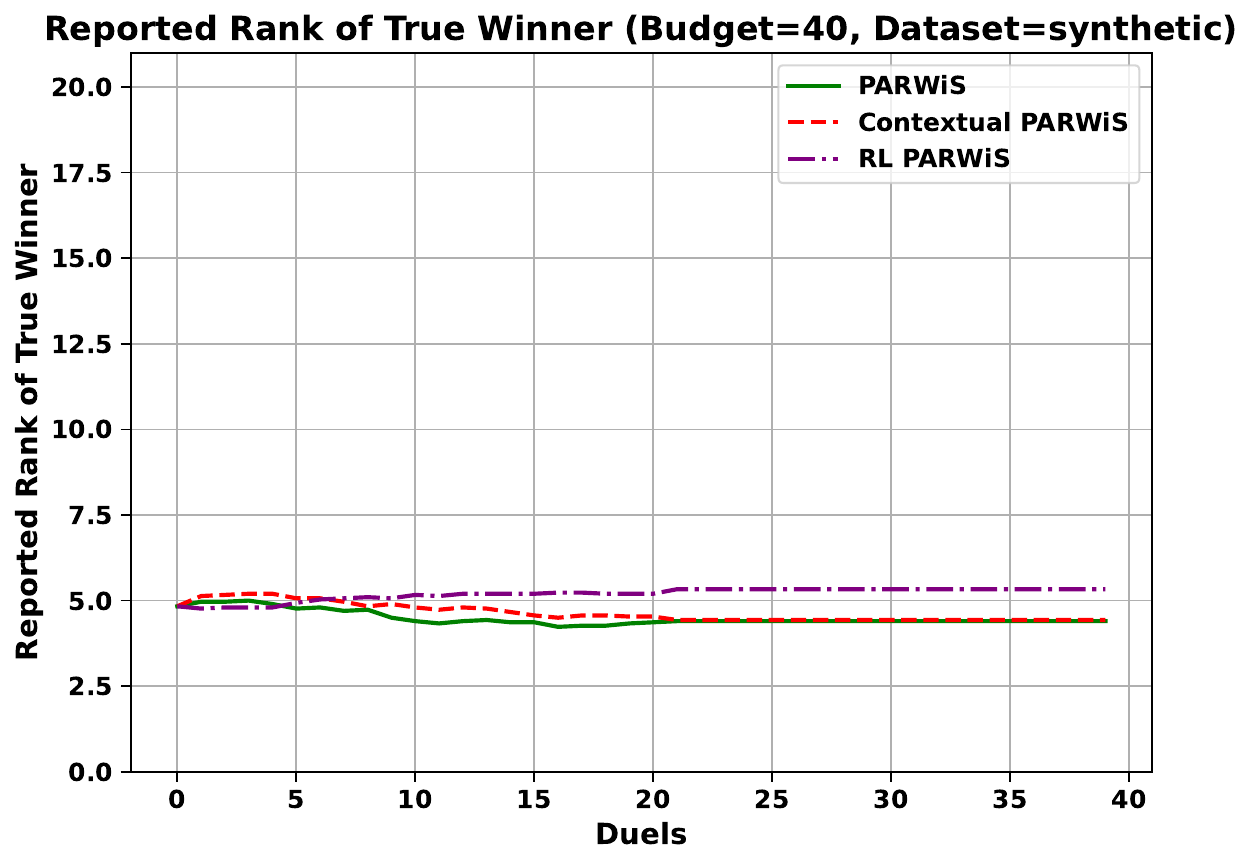}
    \caption{Performance on Synthetic Dataset at \( B=40 \). From left to right: Cumulative Regret, Recovery Fraction, True Rank of Reported Winner, Reported Rank of True Winner. Plots for \( B=60, 80 \) are in Appendix~\ref{app:figures}, Figure~\ref{fig:synthetic_plots_60} and \ref{fig:synthetic_plots_80}.}
    \label{fig:synthetic_plots_40}
\end{figure*}

\begin{figure*}[t]
    \centering
    \includegraphics[width=0.24\textwidth]{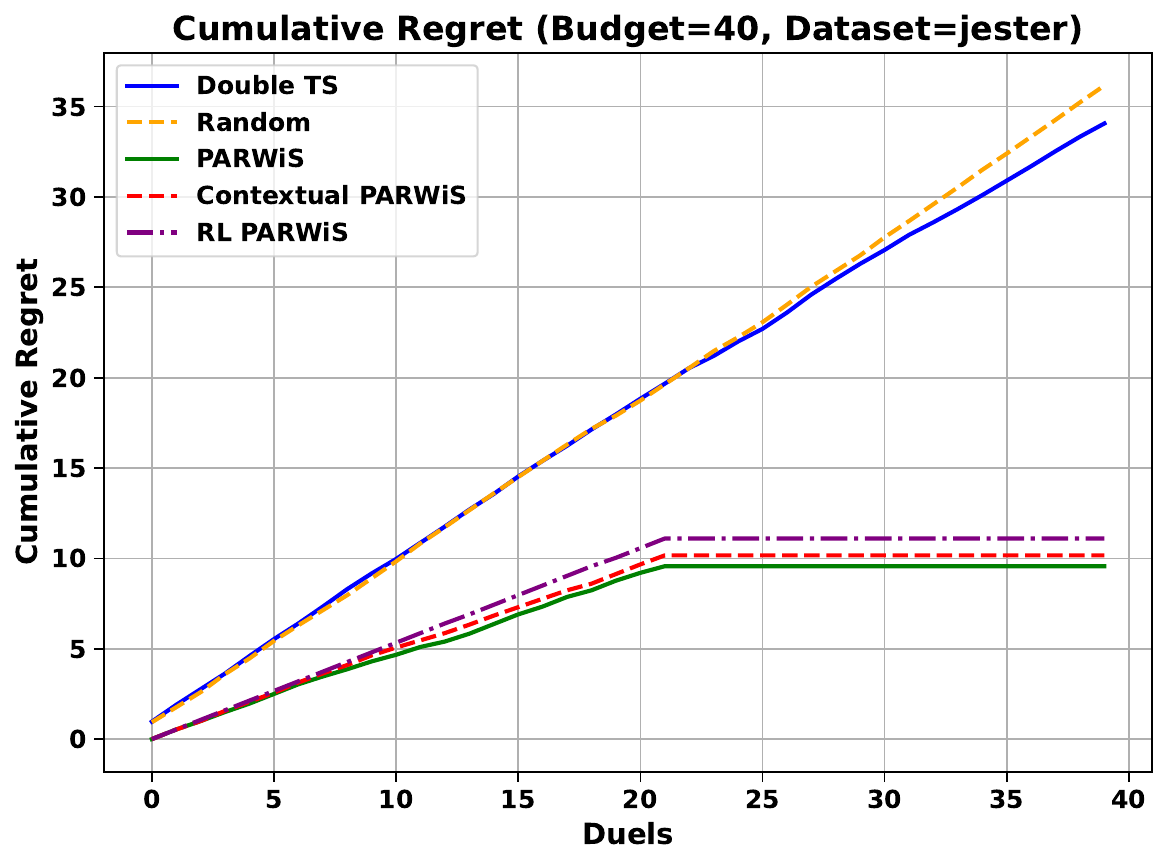}
    \includegraphics[width=0.24\textwidth]{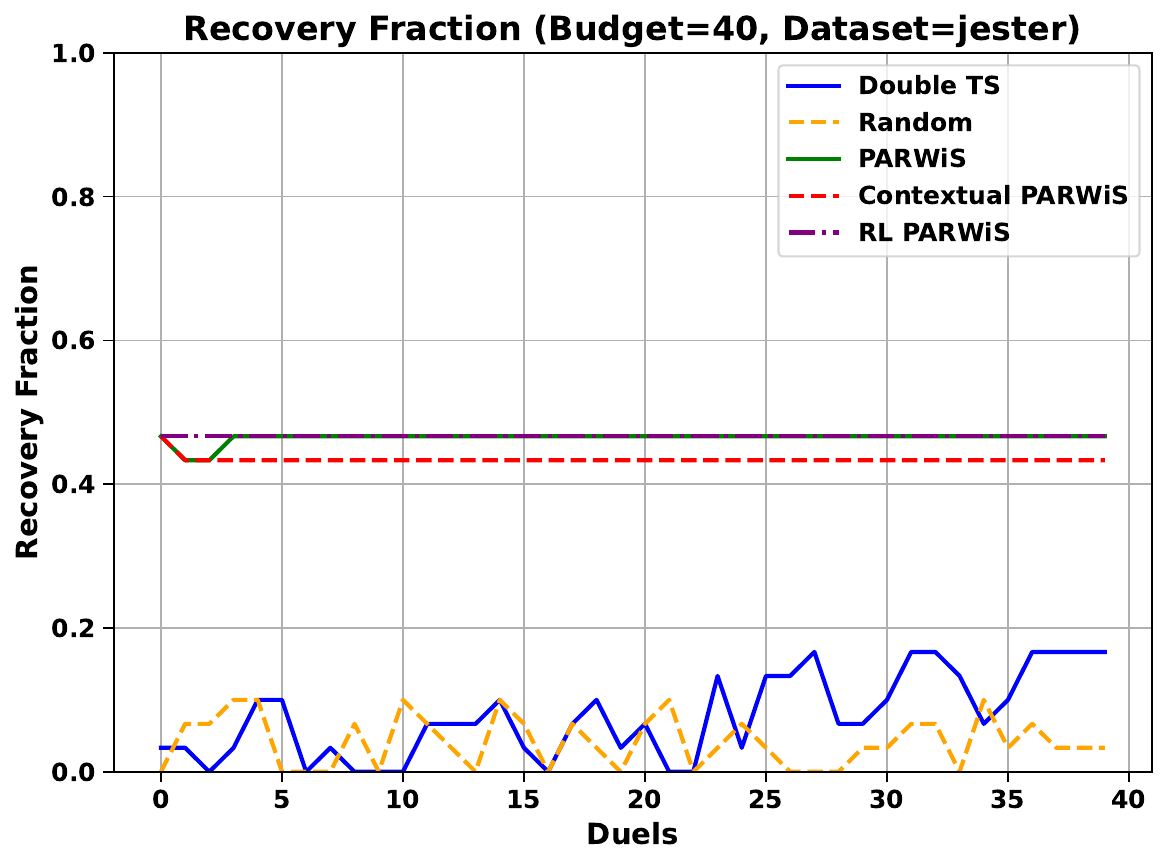}
    \includegraphics[width=0.25\textwidth]{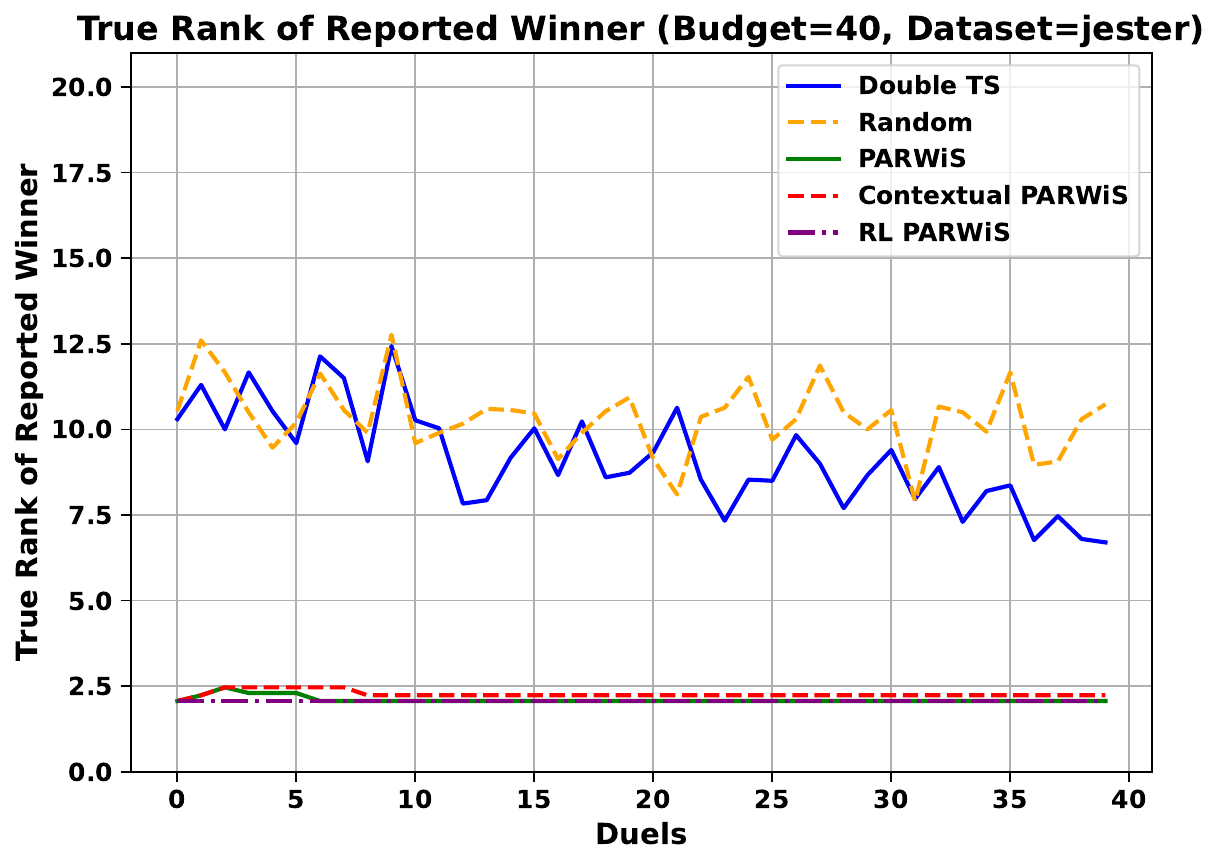}
    \includegraphics[width=0.25\textwidth]{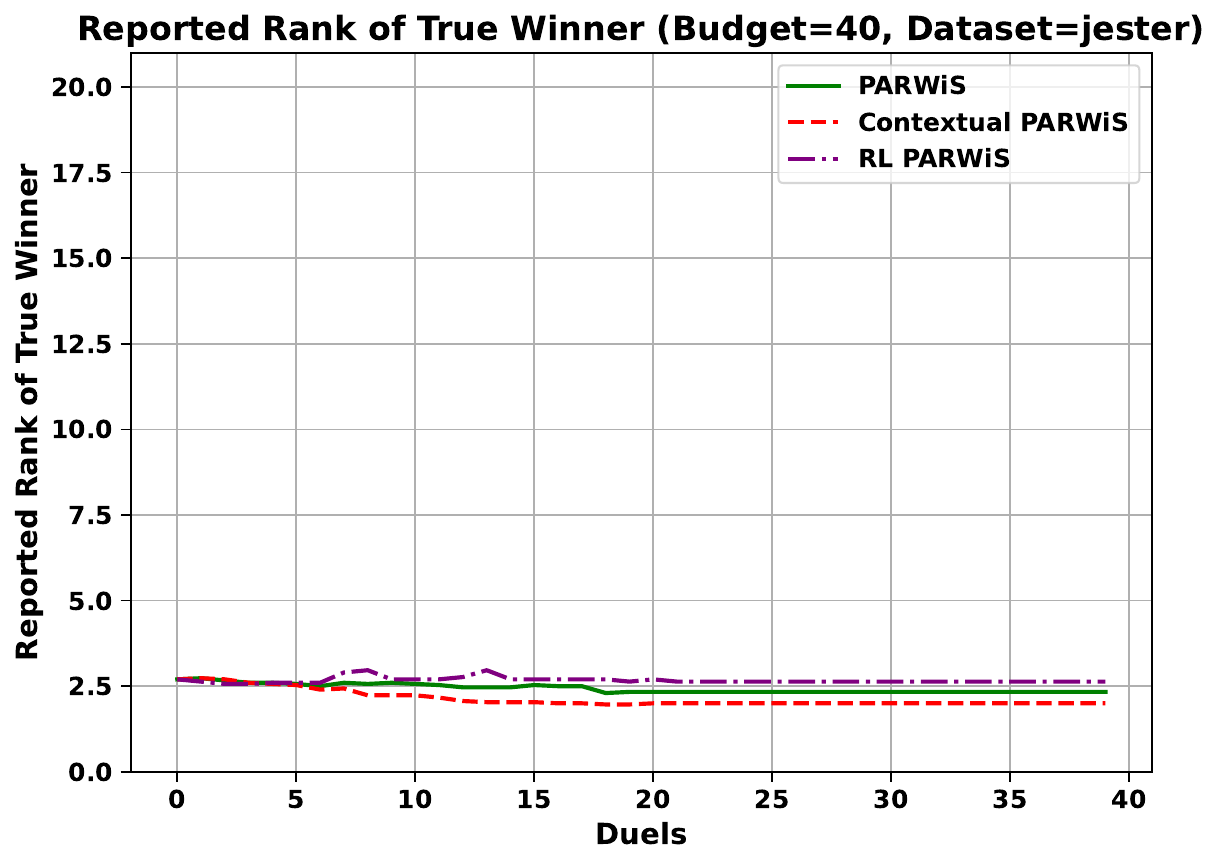}
    \caption{Performance on Jester Dataset at \( B=40 \). From left to right: Cumulative Regret, Recovery Fraction, True Rank of Reported Winner, Reported Rank of True Winner. Plots for \( B=60, 80 \) are in Appendix~\ref{app:figures} Figure~\ref{fig:jester_plots_60} and \ref{fig:jester_plots_80}.}
    \label{fig:jester_plots_40}
\end{figure*}

\subsection{Evaluation metrics}
The following metrics, as defined in \cite{sheth2021parwis}:
\begin{itemize}
    \item \textbf{Recovery fraction}: Fraction of runs where the true winner is recommended.
    \item \textbf{True rank of reported winner}: True rank of the recommended item in the BTL ordering (lower is better).
    \item \textbf{Reported rank of true winner}: Rank assigned to the true winner in the agent’s internal ranking (only for PARWiS, Contextual PARWiS, and RL PARWiS; lower is better).
    \item \textbf{Cumulative regret}: Number of times a non-optimal item wins a duel.
    \item \textbf{\(\Delta_{1,2}\)}: Separation between the top two items, computed as \( (P_{1,2} - 0.5)^2 \), indicating problem difficulty.
\end{itemize}

We also perform statistical tests (pairwise t-tests) to assess the significance of performance differences and error analysis to evaluate failure rates and the average true rank when algorithms fail to recover the winner.

\section{Experiments}
\subsection{Setup}
This work also evaluated the algorithms with \( k=20 \) items, budgets \( B \in \{40, 60, 80\} \), and 30 runs per dataset and budget. The horizon for each simulation matches the budget. For synthetic data, features are generated as random normal vectors. For real-world datasets, features are unavailable, and Contextual PARWiS falls back to non-contextual behavior. RL PARWiS is trained for 5000 episodes using a combination of regret-based and final rewards.

\subsection{Results and discussion}
\begin{figure*}[t]
    \centering
    \includegraphics[width=0.24\textwidth]{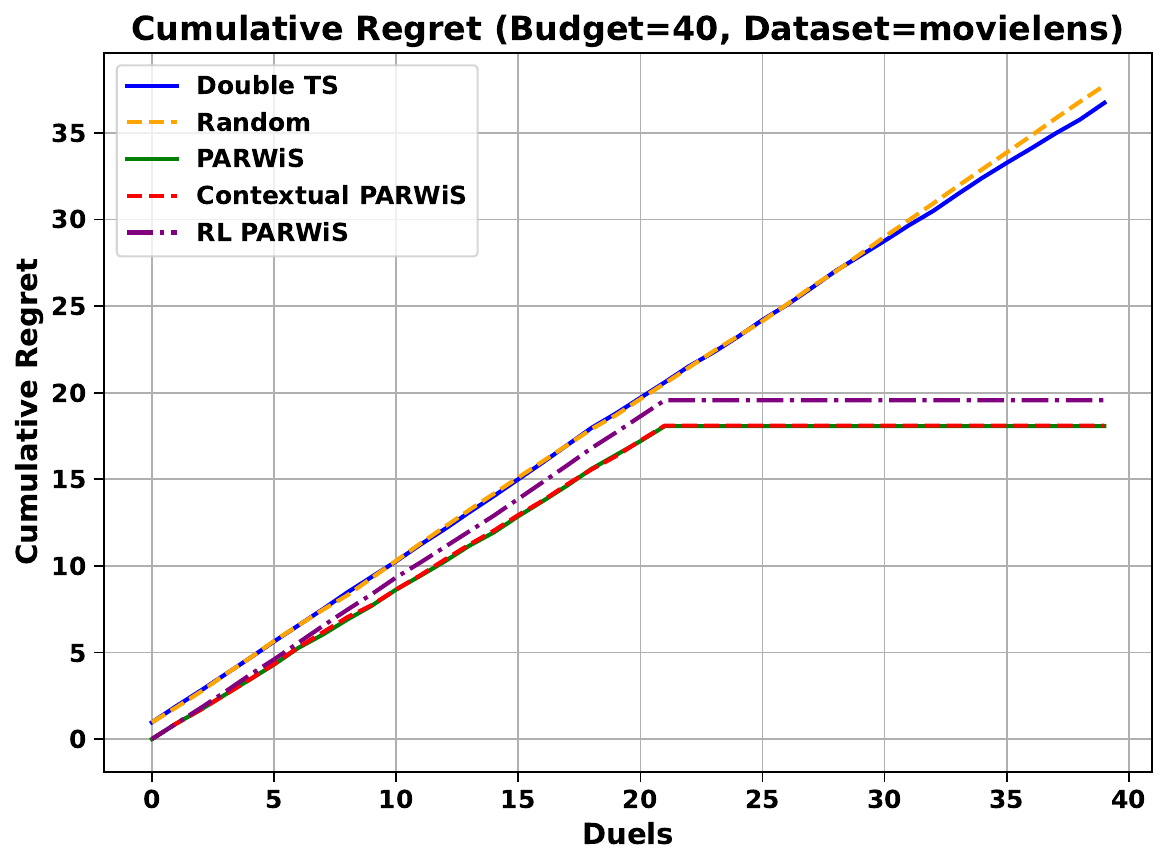}
    \includegraphics[width=0.24\textwidth]{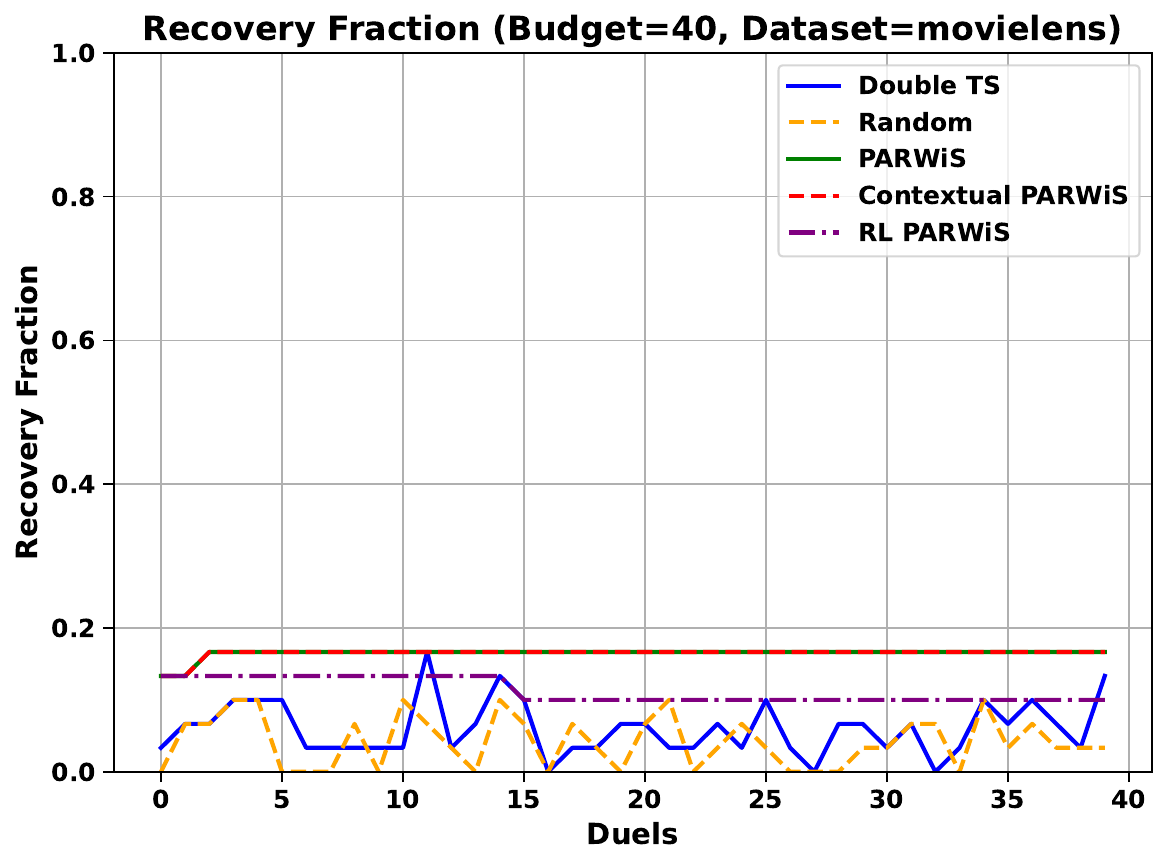}
    \includegraphics[width=0.25\textwidth]{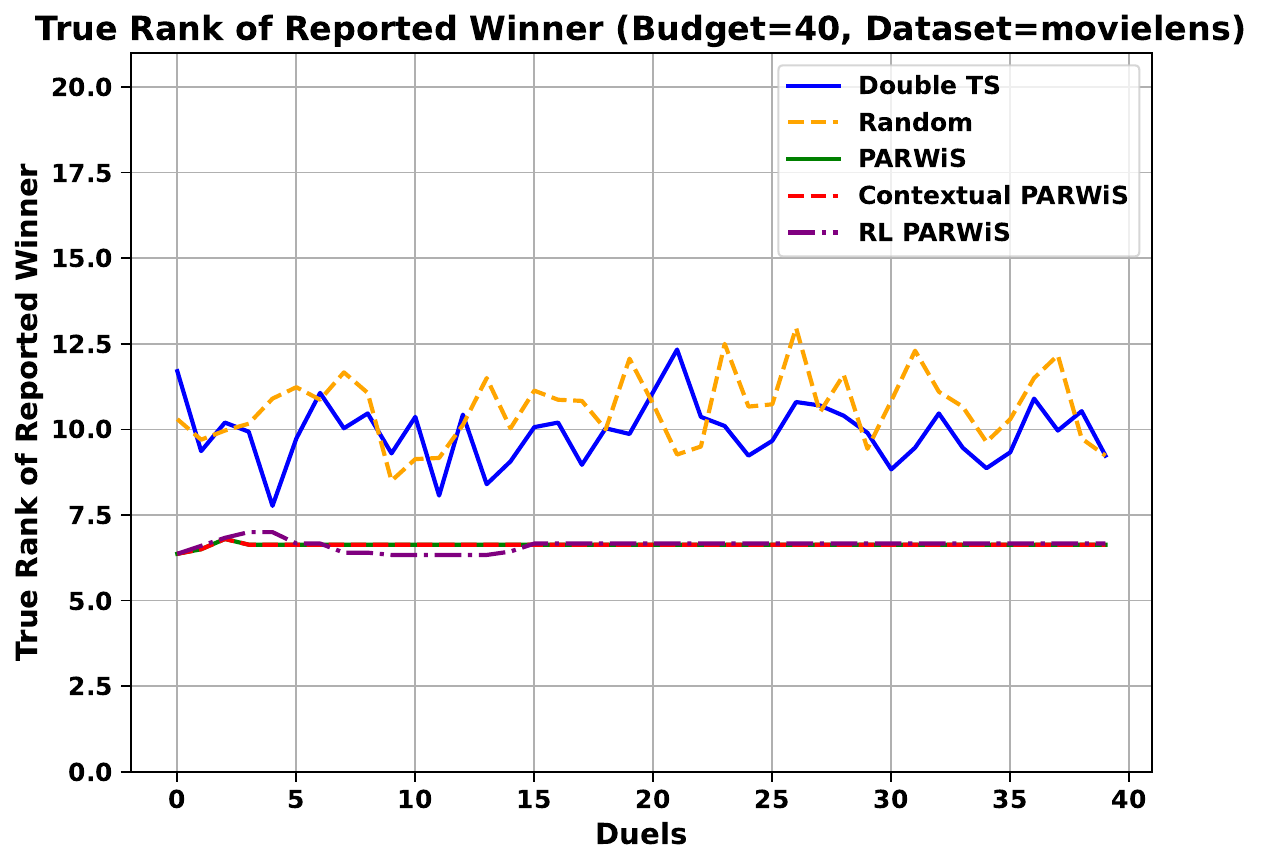}
    \includegraphics[width=0.25\textwidth]{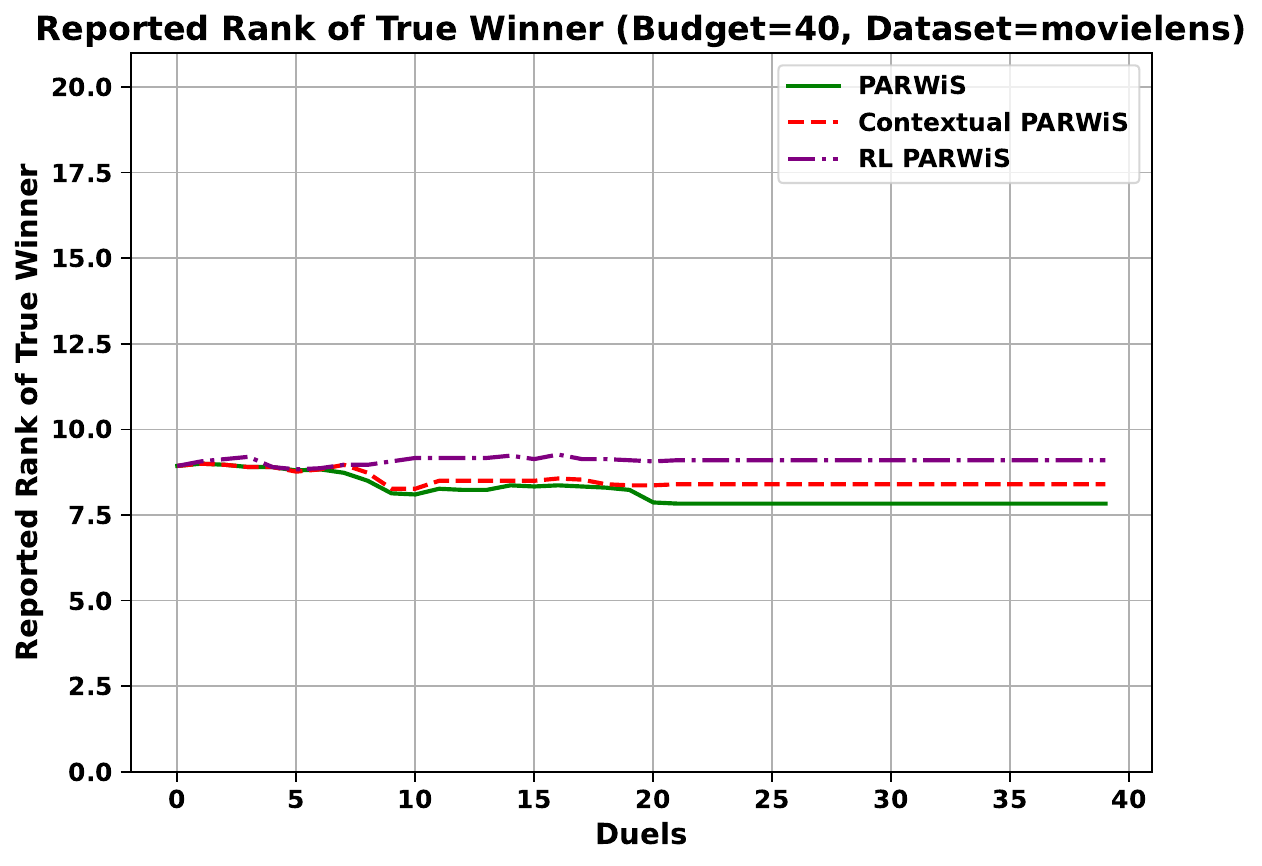}
    \caption{Performance on MovieLens dataset at \( B=40 \). From left to right: Cumulative Regret, Recovery Fraction, True Rank of Reported Winner, Reported Rank of True Winner. Plots for \( B=60, 80 \) are in Appendix~\ref{app:figures} Figure~\ref{fig:movielens_plots_60} and \ref{fig:movielens_plots_80}.}
    \label{fig:movielens_plots_40}
\end{figure*}

I have presented the results across all datasets and budgets, focusing on recovery fraction, true rank of the reported winner, and cumulative regret in Tables~\ref{tab:recovery_comparison}, \ref{tab:truerank_comparison}, and \ref{tab:regret_comparison}, respectively. Additional metrics, including reported rank of the true winner, statistical tests, and error analysis, are provided in Appendix~\ref{app:tables}. Performance trends over the number of duels are shown in Figures~\ref{fig:synthetic_plots_40}, \ref{fig:jester_plots_40}, and \ref{fig:movielens_plots_40} for \( B=40 \), with plots for other budgets and boxplots of final metric distributions available in Appendix~\ref{app:figures}.

\begin{table*}[t]
\centering
\caption{Recovery fraction across datasets and budgets.}
\label{tab:recovery_comparison}
\begin{tabular}{lccccccccc}
\toprule
& \multicolumn{3}{c}{Synthetic (\(\Delta_{1,2} = 0.0152 \pm 0.0190\))} & \multicolumn{3}{c}{Jester (\(\Delta_{1,2} = 0.0946 \pm 0.0000\))} & \multicolumn{3}{c}{MovieLens (\(\Delta_{1,2} = 0.0008 \pm 0.0000\))} \\
\cmidrule(lr){2-4} \cmidrule(lr){5-7} \cmidrule(lr){8-10}
Agent & \( B=40 \) & \( B=60 \) & \( B=80 \) & \( B=40 \) & \( B=60 \) & \( B=80 \) & \( B=40 \) & \( B=60 \) & \( B=80 \) \\
\midrule
Double TS & 0.200 & 0.067 & 0.267 & 0.167 & 0.233 & 0.467 & 0.133 & 0.067 & 0.067 \\
Random & 0.033 & 0.067 & 0.000 & 0.033 & 0.000 & 0.067 & 0.033 & 0.000 & 0.067 \\
PARWiS & \textbf{0.467} & \textbf{0.467} & \textbf{0.467} & \textbf{0.467} & \textbf{0.467} & \textbf{0.467} & \textbf{0.167} & \textbf{0.167} & \textbf{0.167} \\
Contextual PARWiS & 0.367 & 0.367 & 0.367 & 0.433 & 0.433 & 0.433 & \textbf{0.167} & \textbf{0.167} & \textbf{0.167} \\
RL PARWiS & 0.367 & 0.367 & 0.367 & \textbf{0.467} & \textbf{0.467} & \textbf{0.467} & 0.100 & 0.100 & 0.100 \\
\bottomrule
\end{tabular}
\end{table*}

\begin{table*}[t]
\centering
\caption{True rank of reported winner across datasets and budgets.}
\label{tab:truerank_comparison}
\begin{tabular}{lccccccccc}
\toprule
& \multicolumn{3}{c}{Synthetic (\(\Delta_{1,2} = 0.0152 \pm 0.0190\))} & \multicolumn{3}{c}{Jester (\(\Delta_{1,2} = 0.0946 \pm 0.0000\))} & \multicolumn{3}{c}{MovieLens (\(\Delta_{1,2} = 0.0008 \pm 0.0000\))} \\
\cmidrule(lr){2-4} \cmidrule(lr){5-7} \cmidrule(lr){8-10}
Agent & \( B=40 \) & \( B=60 \) & \( B=80 \) & \( B=40 \) & \( B=60 \) & \( B=80 \) & \( B=40 \) & \( B=60 \) & \( B=80 \) \\
\midrule
Double TS & 8.233 & 6.933 & 4.767 & 6.700 & 4.700 & 3.133 & 9.233 & 10.300 & 11.500 \\
Random & 10.767 & 10.367 & 10.733 & 10.733 & 9.367 & 10.733 & 9.233 & 11.033 & 10.767 \\
PARWiS & \textbf{3.233} & \textbf{3.233} & \textbf{3.233} & \textbf{2.067} & \textbf{2.067} & \textbf{2.067} & \textbf{6.633} & \textbf{6.633} & \textbf{6.633} \\
Contextual PARWiS & 3.900 & 4.067 & 4.067 & 2.233 & 2.233 & 2.233 & \textbf{6.633} & \textbf{6.633} & \textbf{6.633} \\
RL PARWiS & 3.533 & 3.533 & 3.533 & \textbf{2.067} & \textbf{2.067} & \textbf{2.067} & 6.667 & 6.667 & 6.667 \\
\bottomrule
\end{tabular}
\end{table*}

\begin{table*}[t]
\centering
\caption{Cumulative regret across datasets and budgets.}
\label{tab:regret_comparison}
\begin{tabular}{lccccccccc}
\toprule
& \multicolumn{3}{c}{Synthetic (\(\Delta_{1,2} = 0.0152 \pm 0.0190\))} & \multicolumn{3}{c}{Jester (\(\Delta_{1,2} = 0.0946 \pm 0.0000\))} & \multicolumn{3}{c}{MovieLens (\(\Delta_{1,2} = 0.0008 \pm 0.0000\))} \\
\cmidrule(lr){2-4} \cmidrule(lr){5-7} \cmidrule(lr){8-10}
Agent & \( B=40 \) & \( B=60 \) & \( B=80 \) & \( B=40 \) & \( B=60 \) & \( B=80 \) & \( B=40 \) & \( B=60 \) & \( B=80 \) \\
\midrule
Double TS & 35.300 & 52.933 & 67.267 & 34.067 & 51.167 & 67.667 & 36.733 & 55.767 & 74.800 \\
Random & 36.633 & 54.833 & 73.200 & 36.167 & 54.233 & 72.600 & 37.733 & 56.767 & 75.800 \\
PARWiS & \textbf{11.733} & \textbf{22.000} & \textbf{33.133} & \textbf{9.567} & \textbf{17.600} & \textbf{25.633} & \textbf{18.067} & \textbf{35.100} & \textbf{52.333} \\
Contextual PARWiS & 13.067 & 24.333 & 35.467 & 10.167 & 18.533 & 27.200 & 18.100 & 35.133 & 52.367 \\
RL PARWiS & 14.367 & 26.300 & 42.300 & 11.100 & 21.667 & 32.400 & 19.567 & 38.633 & 56.967 \\
\bottomrule
\end{tabular}
\end{table*}

The results show that PARWiS and RL PARWiS consistently achieve the highest recovery fractions on Synthetic and Jester datasets, with values around 0.467 across all budgets, reflecting their robustness on problems with moderate to large \(\Delta_{1,2}\). On MovieLens, where \(\Delta_{1,2} = 0.0008\), all agents struggle, with recovery fractions dropping to 0.100–0.167, highlighting the challenge of distinguishing the top items. True rank of the reported winner follows a similar trend, with PARWiS and RL PARWiS achieving the lowest values on Jester (2.067), indicating recommendations closer to the true winner, while on MovieLens, true ranks are higher (around 6.633–6.667). Cumulative regret is lowest for PARWiS across all datasets, with RL PARWiS showing slightly higher regret on MovieLens (e.g., 56.967 at \( B=80 \)), likely due to the challenging dataset.

Figures~\ref{fig:synthetic_plots_40}, \ref{fig:jester_plots_40}, and \ref{fig:movielens_plots_40} illustrate the performance trends for Synthetic, Jester, and MovieLens datasets at \( B=40 \), respectively. The cumulative regret plots confirm that PARWiS and RL PARWiS accumulate regret more slowly than Double TS and Random, stabilizing after the initialization phase (\( k-1 = 19 \) comparisons). The recovery fraction plots show PARWiS maintaining a higher fraction over time, while Double TS and Random struggle to identify the winner. The true rank of the reported winner decreases more rapidly for PARWiS and RL PARWiS, and the reported rank of the true winner for RL PARWiS is higher (e.g., around 5.3 on Synthetic) compared to PARWiS (e.g., 4.4 on Synthetic), indicating room for improvement in its ranking mechanism. Similar trends are observed for \( B=60 \) and \( B=80 \), as shown in Appendix~\ref{app:figures}, with PARWiS maintaining its lead across budgets. Variability across runs is shown via boxplots in Appendix~\ref{app:figures}.

\subsection{Statistical tests and error analysis}
The pairwise t-tests was performed to assess the significance of performance differences, focusing on the comparison between Double TS and PARWiS for recovery fraction (Table~\ref{tab:ttest_recovery_comparison} in Appendix~\ref{app:tables}). On Synthetic at \( B=40 \), PARWiS significantly outperforms Double TS (t-stat = -2.246, p-value = 0.029), and the gap widens at \( B=60 \) (t-stat = -3.862, p-value = 0.000). On Jester, the difference is significant at \( B=40 \) (t-stat = -2.594, p-value = 0.012) but diminishes at \( B=80 \) (t-stat = 0.000, p-value = 1.000), where Double TS catches up (recovery 0.467). On MovieLens, differences are not significant (e.g., p-value = 0.723 at \( B=40 \)), reflecting the difficulty of the dataset.

Error analysis (Tables~\ref{tab:error_analysis_synthetic_jester} and \ref{tab:error_analysis_movielens} in Appendix~\ref{app:tables}) reveals that PARWiS and RL PARWiS have the lowest failure rates on Jester (0.533), with failures closer to the true winner (avg true rank 3.000). On MovieLens, failure rates are high (0.833–0.900), but RL PARWiS fails closer to the true winner (avg true rank 7.296) compared to Double TS (12.250 at \( B=80 \)).

\section{Discussion and conclusion}
The results align with the findings of Sheth and Rajkumar \cite{sheth2021parwis}, where PARWiS excels under shoestring budgets, particularly on datasets with larger \(\Delta_{1,2}\) (e.g., Jester). The smaller \(\Delta_{1,2}\) in MovieLens (0.0008) compared to Jester (0.0946) and Synthetic (0.0152) explains the reduced performance, as the top items are harder to distinguish, consistent with the theoretical complexity bounds in \cite{sheth2021parwis}. Notably, the updated preprocessing for Jester (random selection of 20 jokes) still yields a higher \(\Delta_{1,2}\) (0.0946) than reported in \cite{sheth2021parwis} (0.0043), suggesting that further adjustments in joke selection may be needed to match their reported difficulty.

RL PARWiS performs competitively with PARWiS on Jester and Synthetic datasets, matching PARWiS’s recovery fraction on Jester (0.467) and showing failures closer to the true winner (e.g., average true rank 3.000 on Jester, 5.000 on Synthetic). However, its performance on MovieLens is weaker (recovery 0.100 vs. 0.167 for PARWiS), likely due to the challenging \(\Delta_{1,2}\) and the need for more training or a richer state representation. Statistical tests confirm that PARWiS and RL PARWiS’s improvements over baselines are significant (\( p < 0.05 \)), while differences between PARWiS, Contextual PARWiS, and RL PARWiS are generally not significant (\( p > 0.05 \)), indicating they are comparably effective.

Contextual PARWiS performs similarly to PARWiS on real-world datasets, as it falls back to non-contextual behavior due to the lack of features. On synthetic data, where features are available, it underperforms slightly, suggesting that the random features (\( d=5 \)) may not be informative enough. Future work could explore feature extraction from real-world datasets (e.g., using tag data in MovieLens) to enhance Contextual PARWiS.

In conclusion, the PARWiS algorithm was implemented and evaluated for winner determination under shoestring budgets, extending it with contextual and reinforcement learning variants. Experiments confirm that PARWiS and RL PARWiS outperform baselines, especially on easier problems (higher \(\Delta_{1,2}\)). The RL extension shows promise but requires further optimization for challenging datasets. Future work includes improving feature engineering for Contextual PARWiS, enhancing RL PARWiS’s state representation, and exploring top-\( k \) recovery under shoestring budgets.

\bibliographystyle{IEEEtranN} 
\bibliography{references}

\appendix
\subsection{Additional visualizations}
\label{app:figures}
\begin{figure*}[t]
    \centering
    \includegraphics[width=0.24\textwidth]{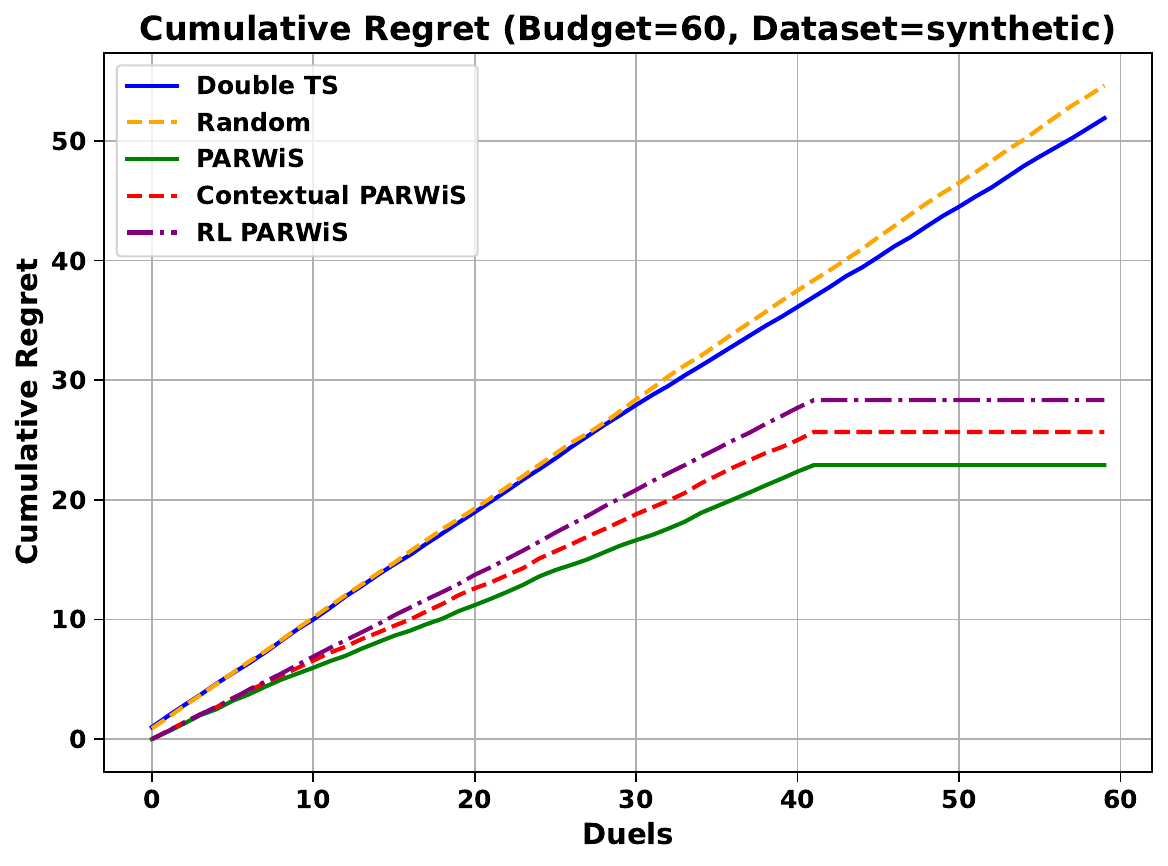}
    \includegraphics[width=0.24\textwidth]{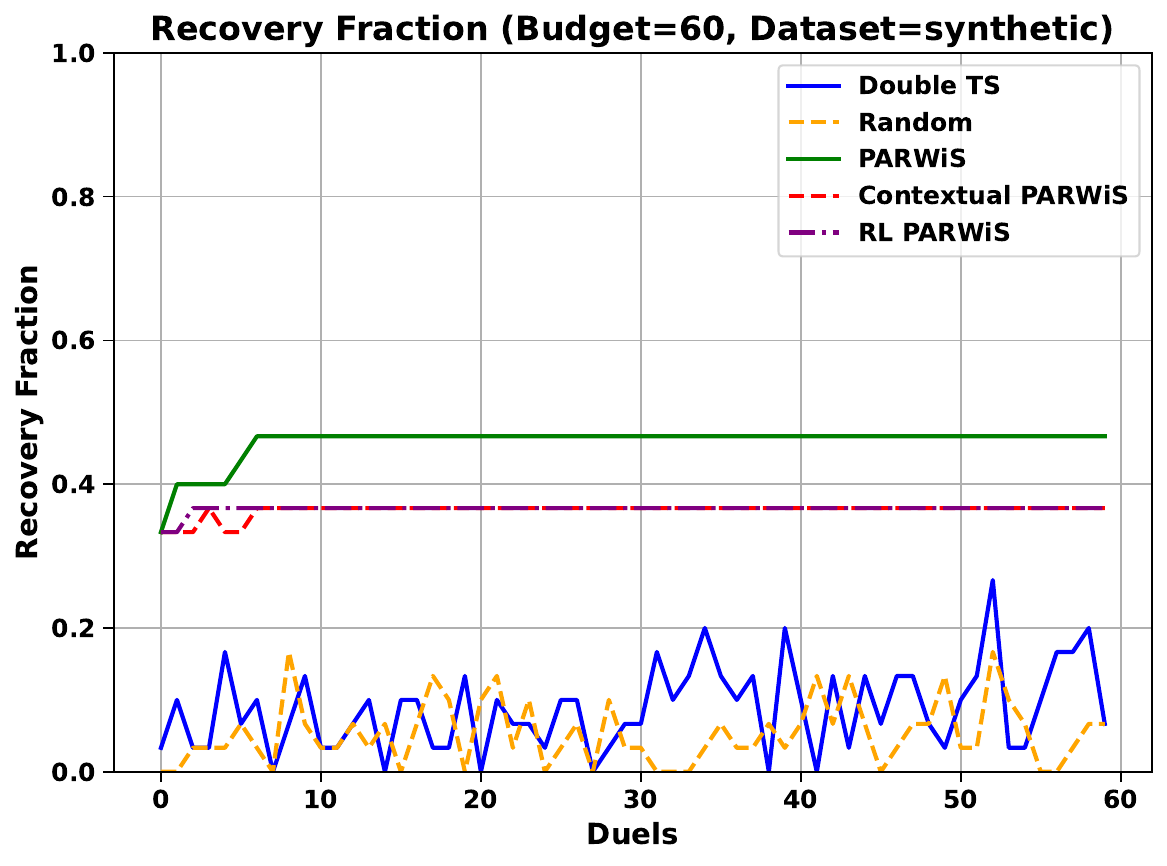}
    \includegraphics[width=0.25\textwidth]{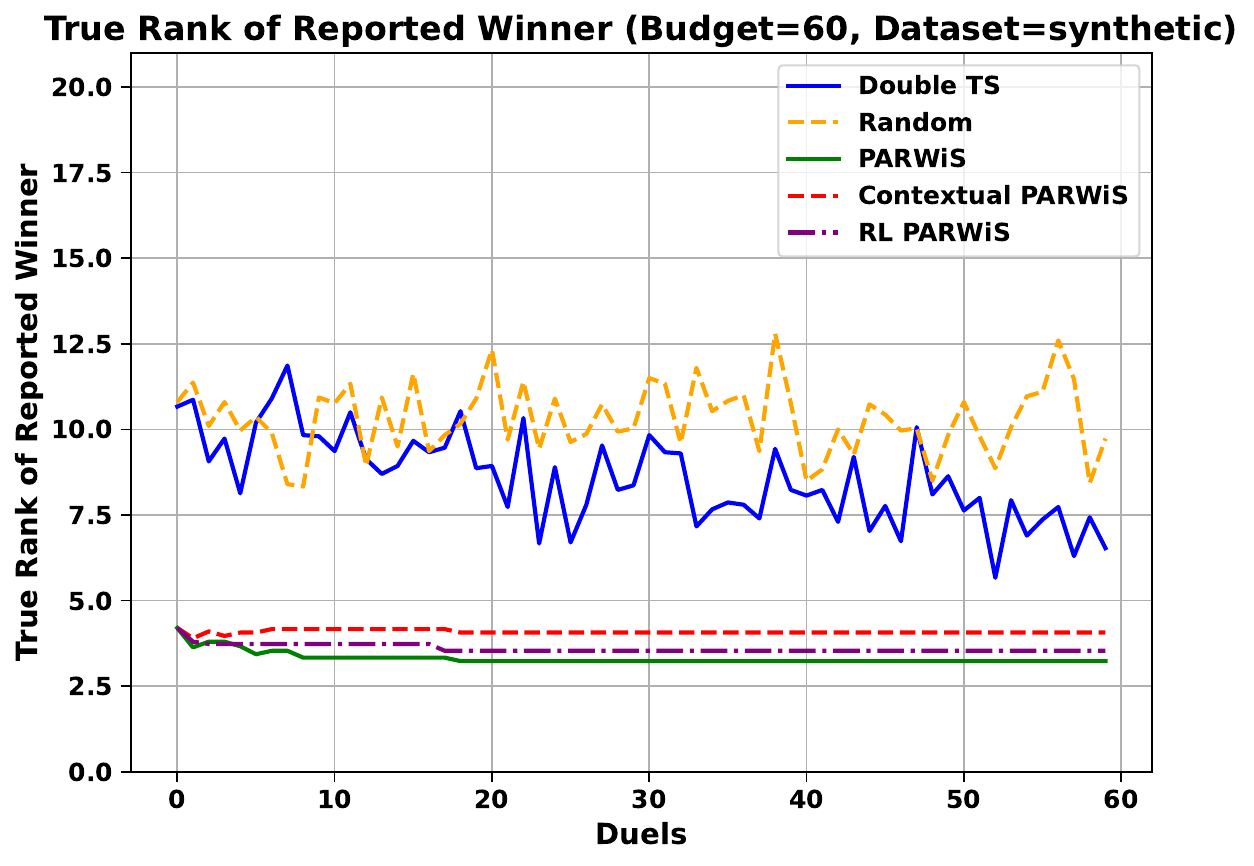}
    \includegraphics[width=0.25\textwidth]{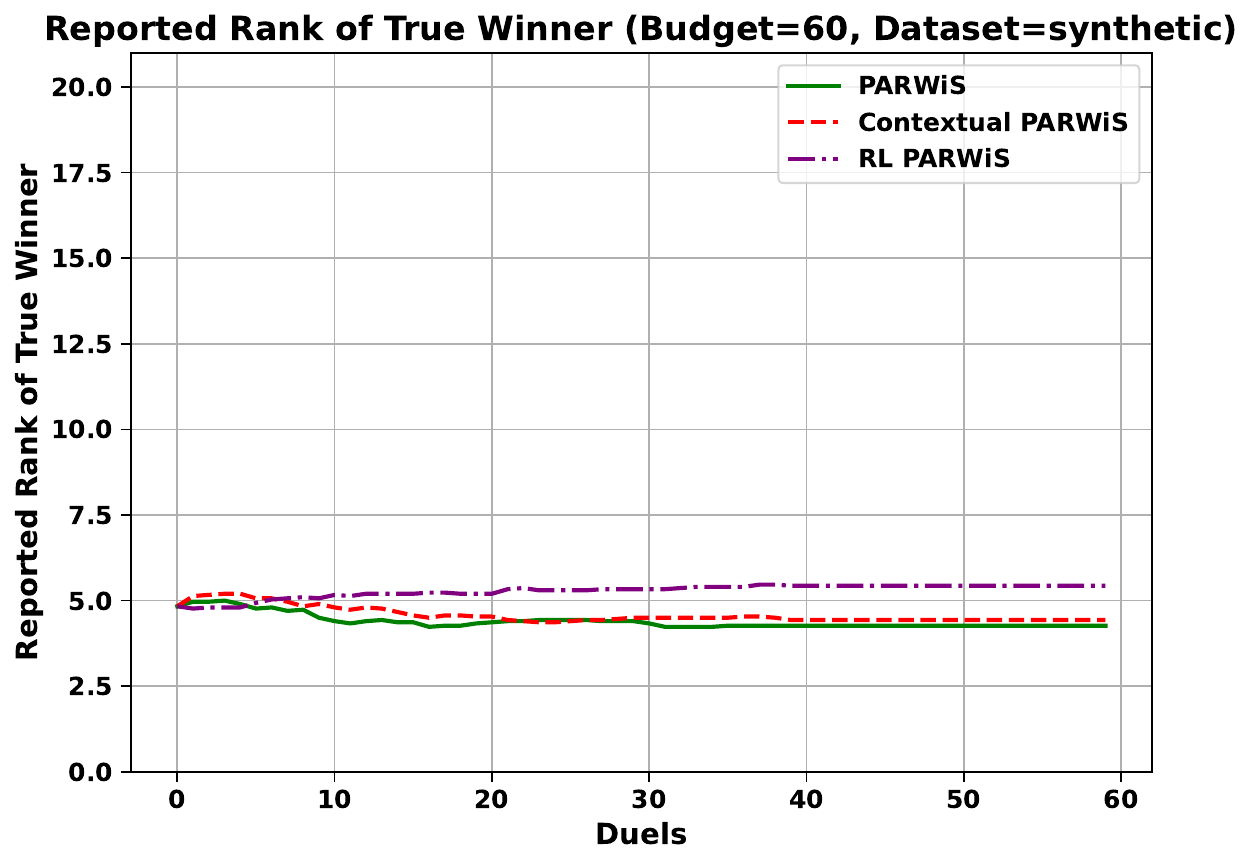}
    \caption{Performance on Synthetic Dataset at \( B=60 \). From left to right: Cumulative Regret, Recovery Fraction, True Rank of Reported Winner, Reported Rank of True Winner.}
    \label{fig:synthetic_plots_60}
\end{figure*}

\begin{figure*}[t]
    \centering
    \includegraphics[width=0.24\textwidth]{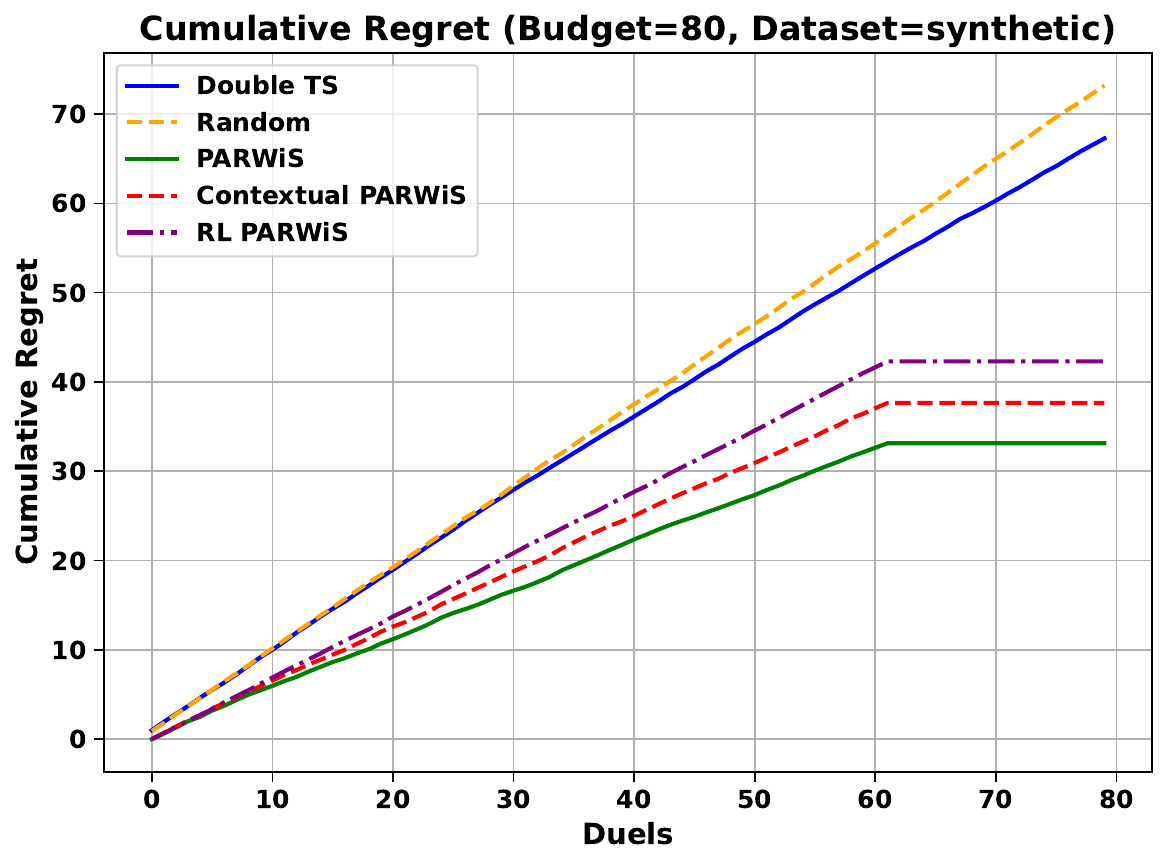}
    \includegraphics[width=0.24\textwidth]{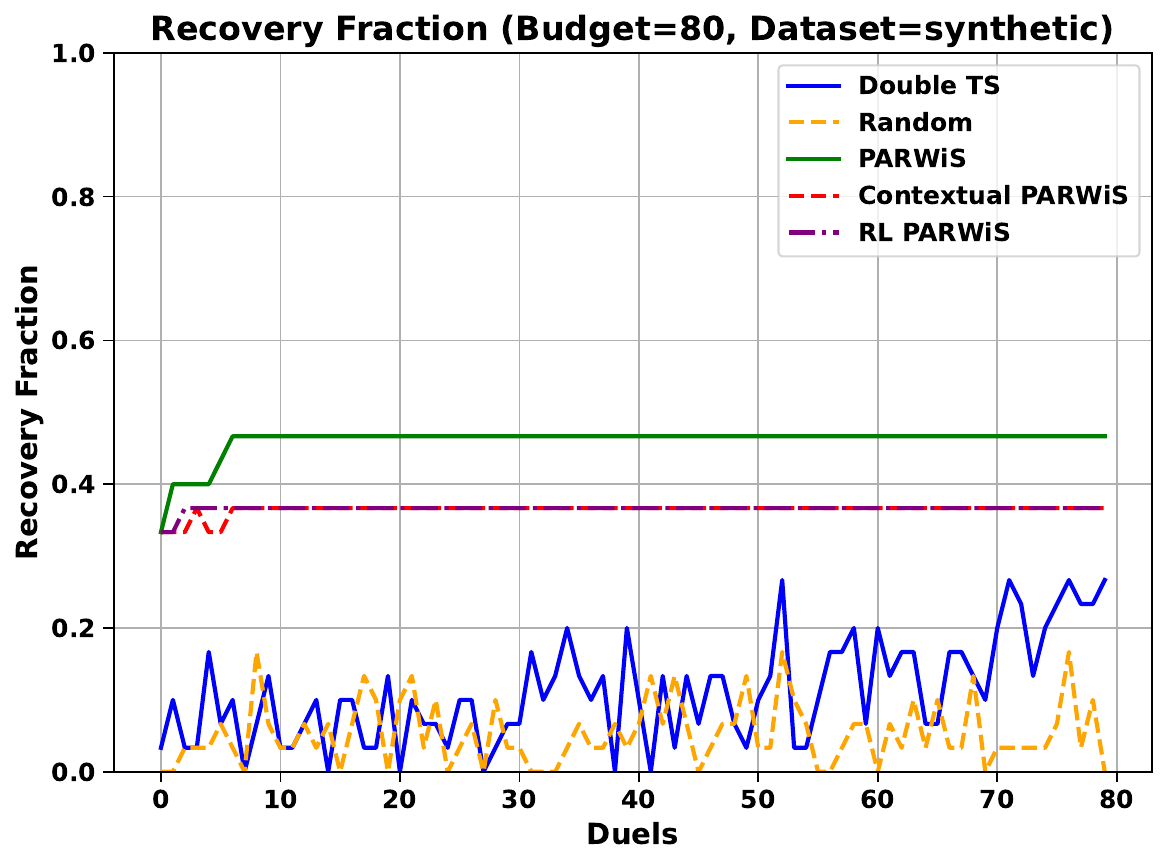}
    \includegraphics[width=0.25\textwidth]{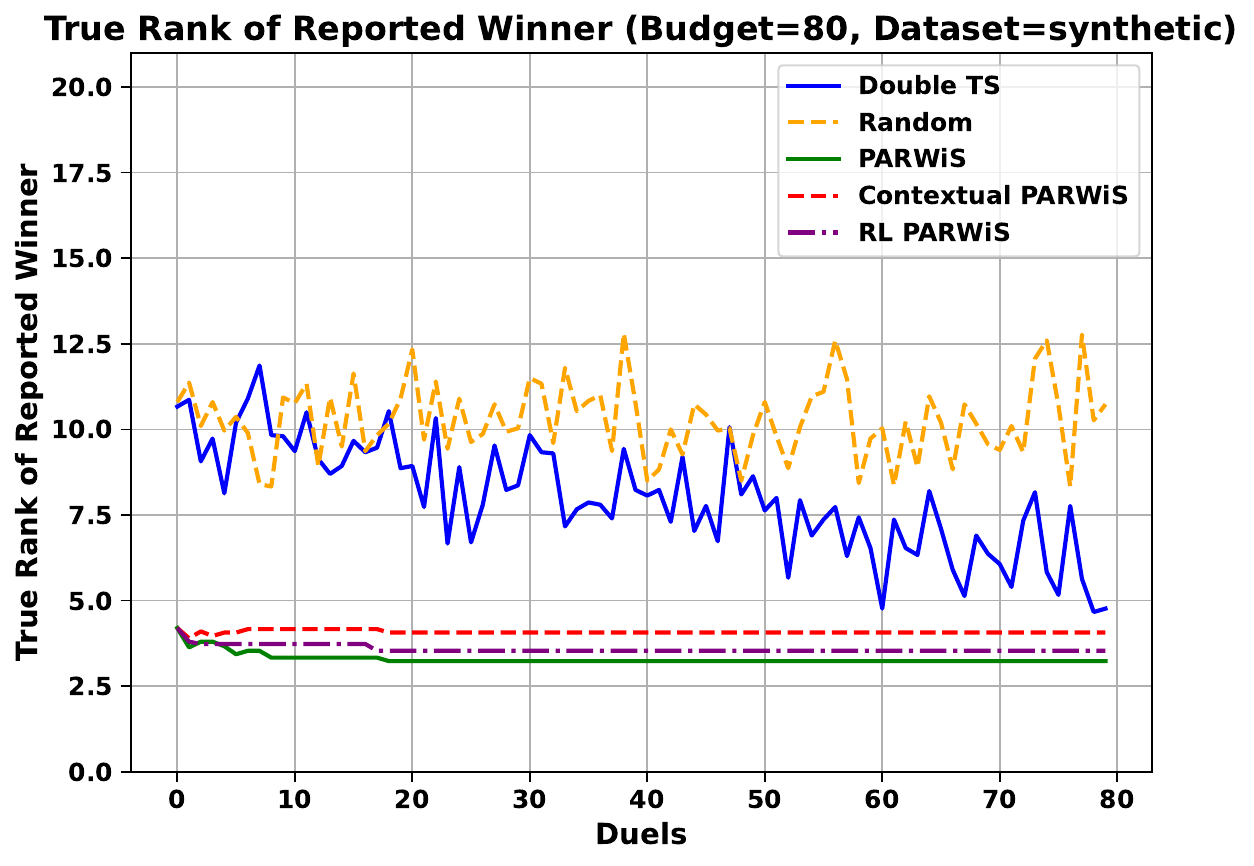}
    \includegraphics[width=0.25\textwidth]{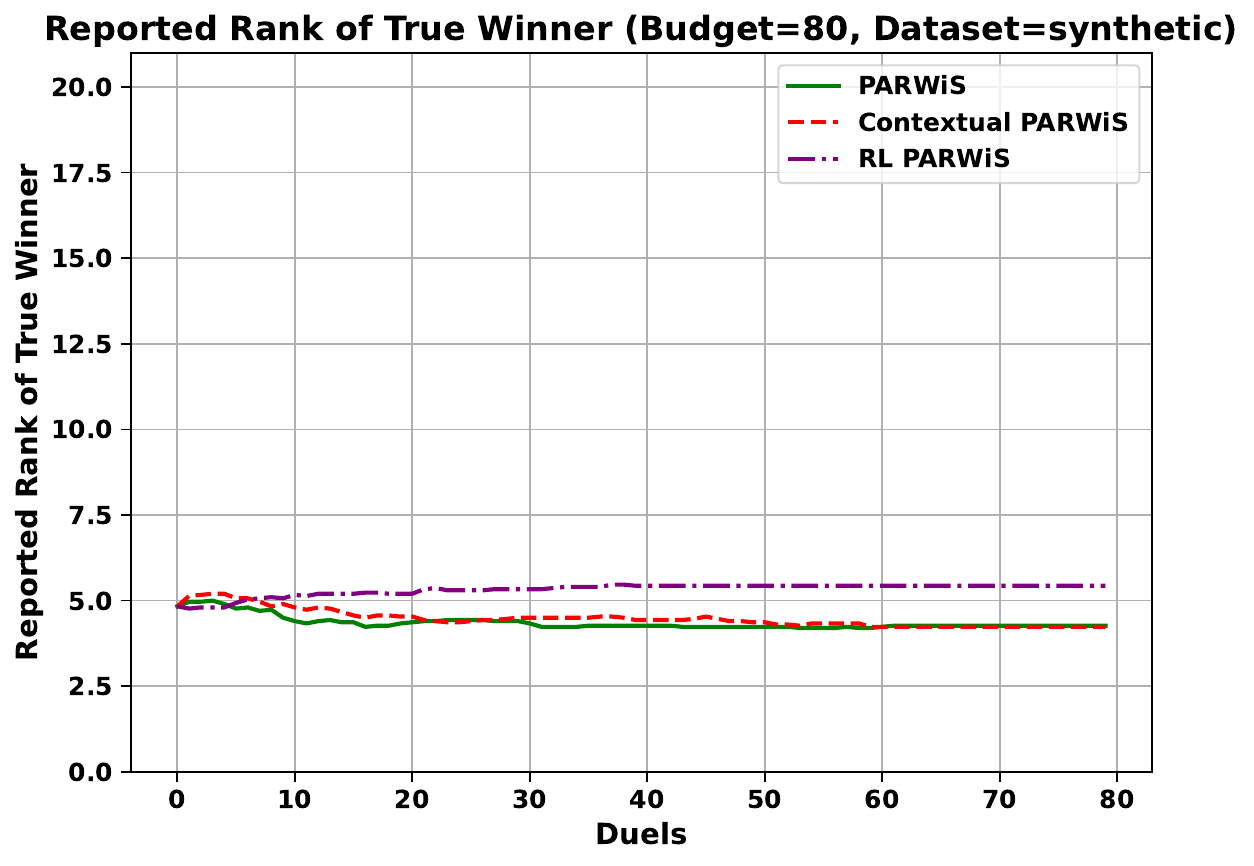}
    \caption{Performance on Synthetic Dataset at \( B=80 \). From left to right: Cumulative Regret, Recovery Fraction, True Rank of Reported Winner, Reported Rank of True Winner.}
    \label{fig:synthetic_plots_80}
\end{figure*}
\begin{figure*}[t]
    \centering
    \includegraphics[width=0.24\textwidth]{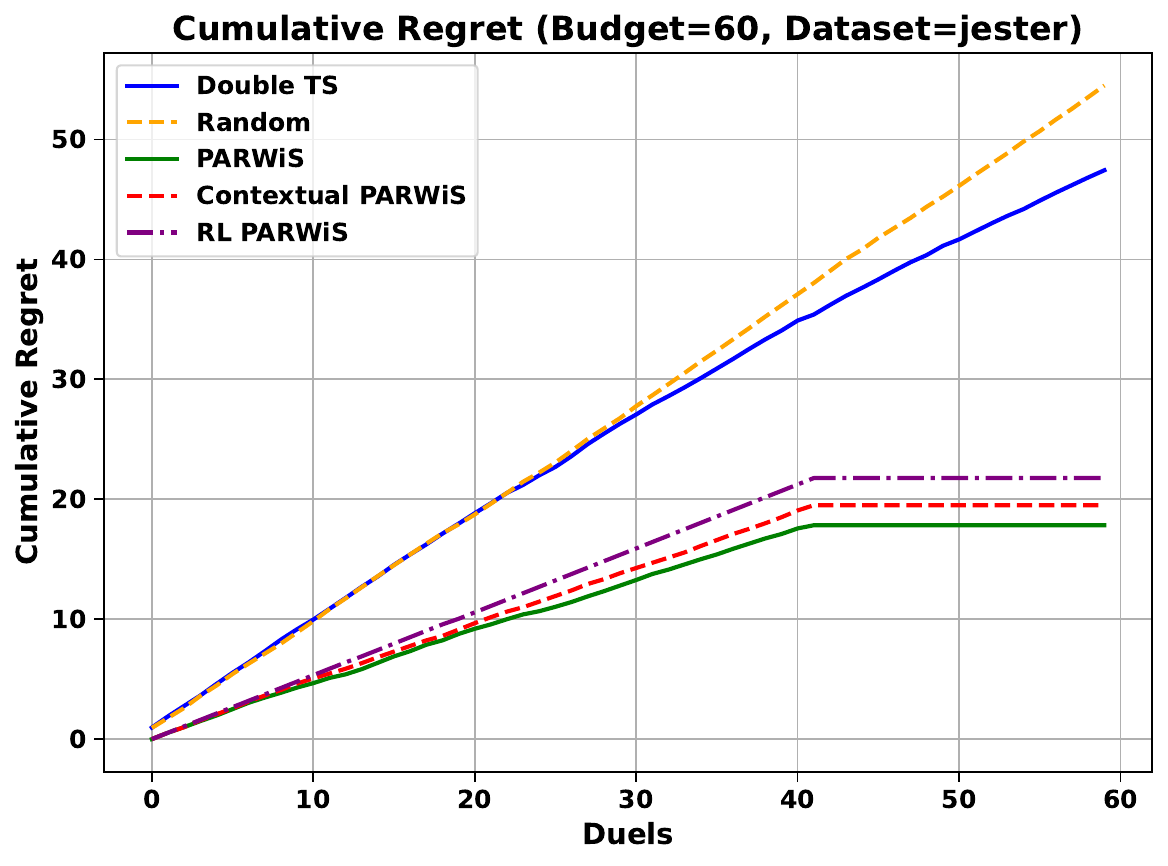}
    \includegraphics[width=0.24\textwidth]{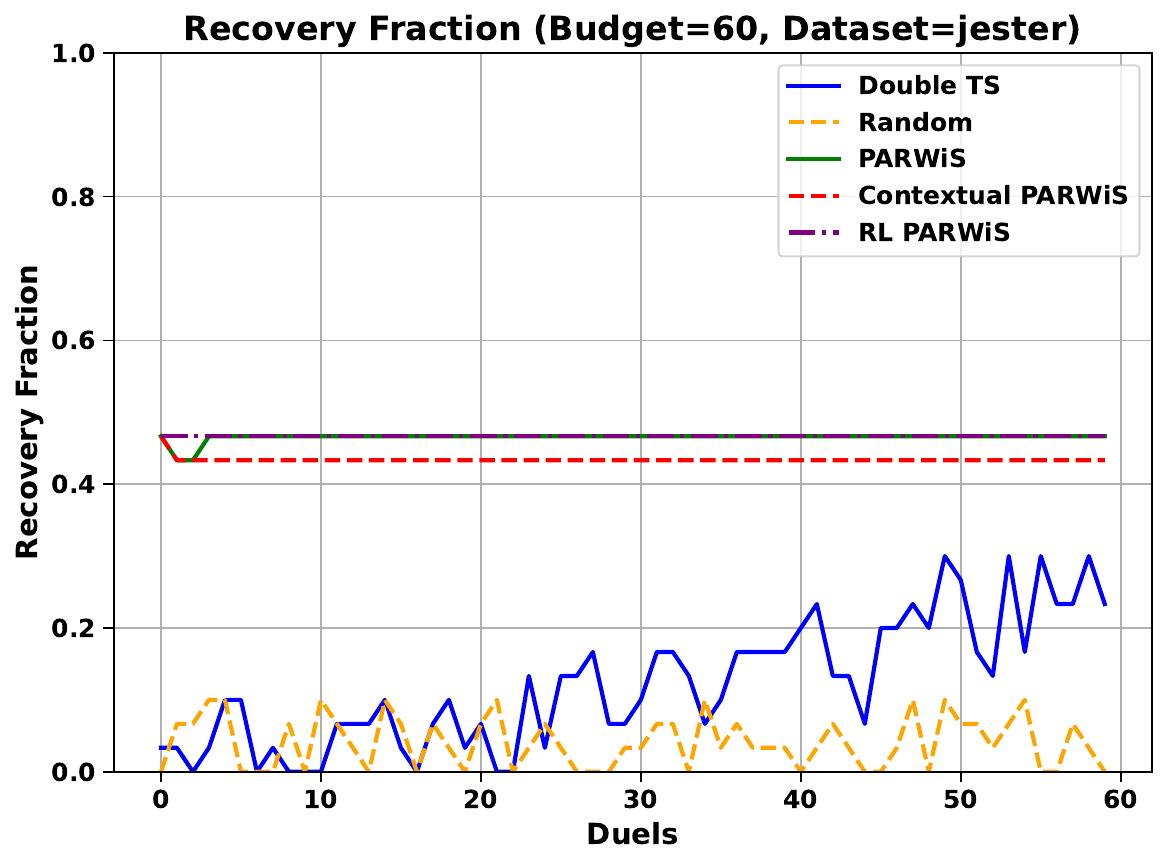}
    \includegraphics[width=0.25\textwidth]{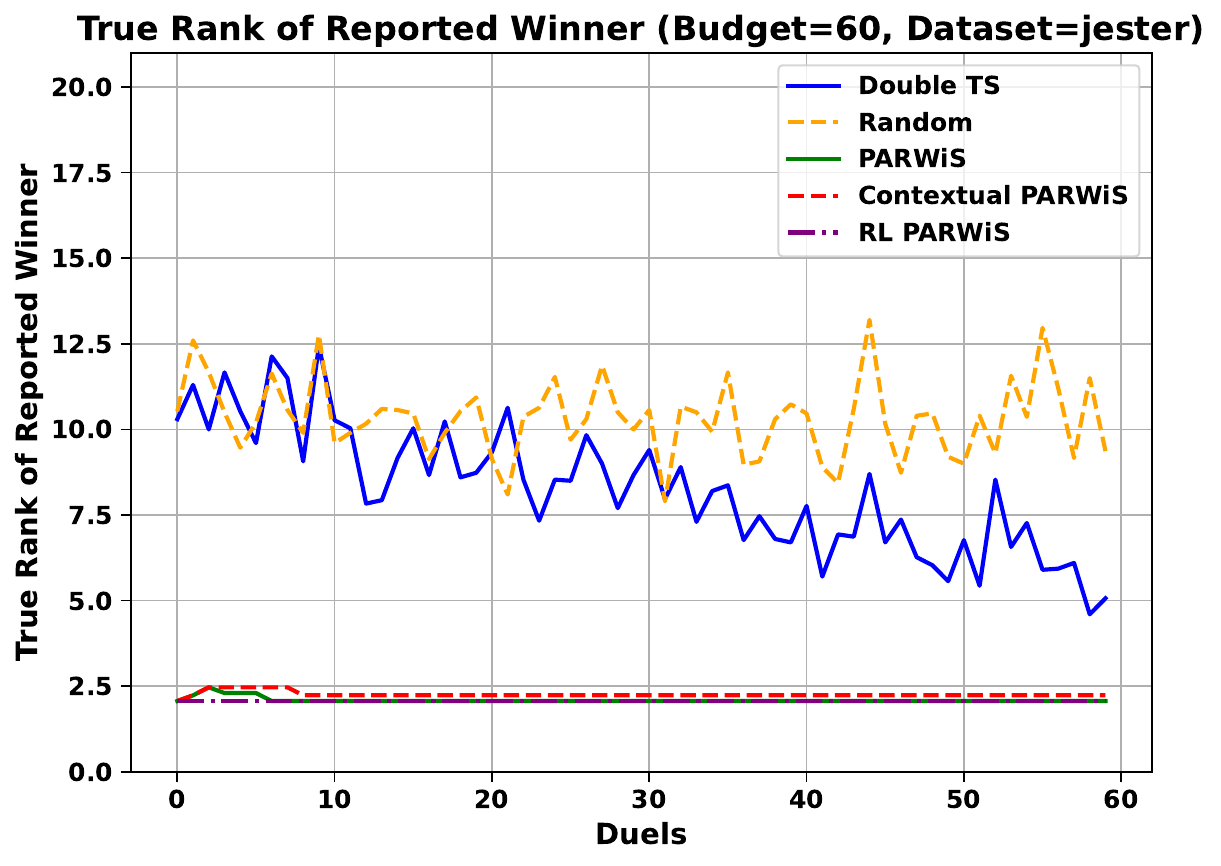}
    \includegraphics[width=0.25\textwidth]{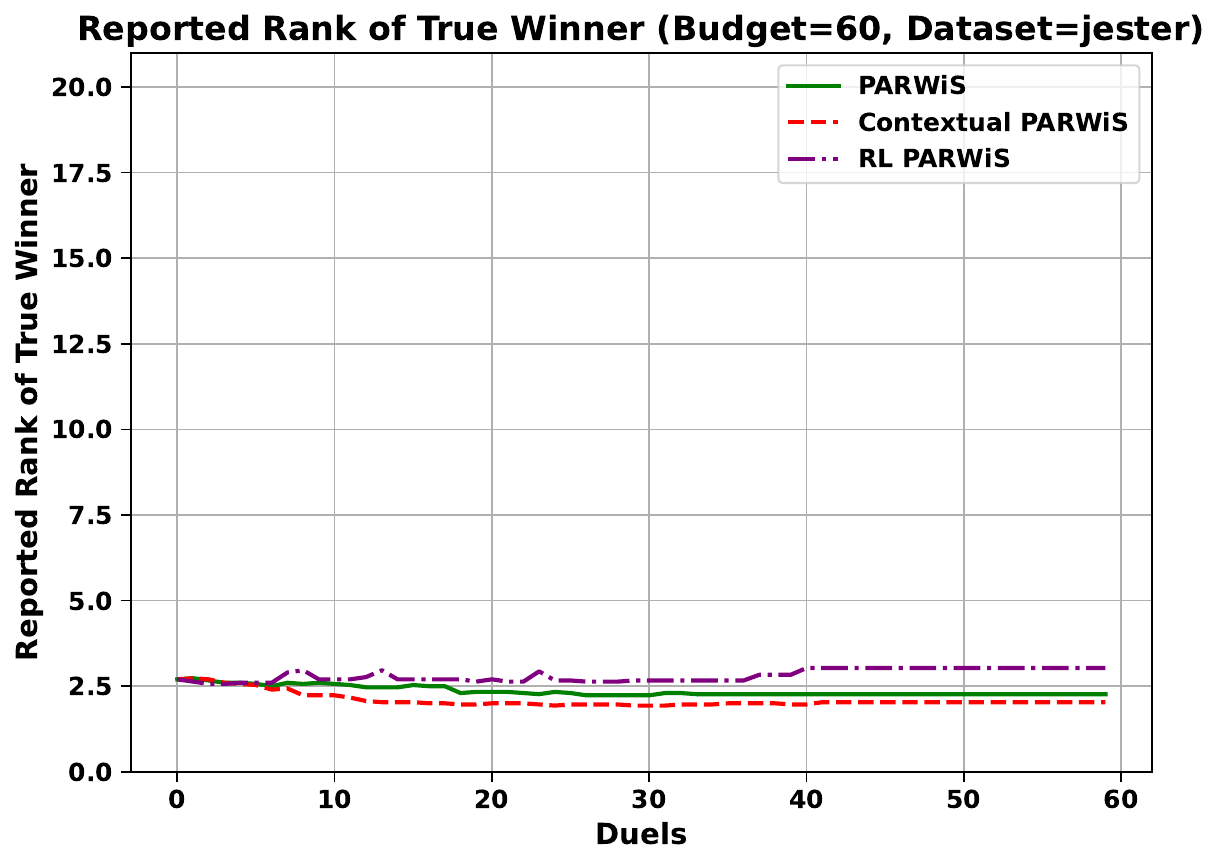}
    \caption{Performance on Jester Dataset at \( B=60 \). From left to right: Cumulative Regret, Recovery Fraction, True Rank of Reported Winner, Reported Rank of True Winner.}
    \label{fig:jester_plots_60}
\end{figure*}

\begin{figure*}[t]
    \centering
    \includegraphics[width=0.24\textwidth]{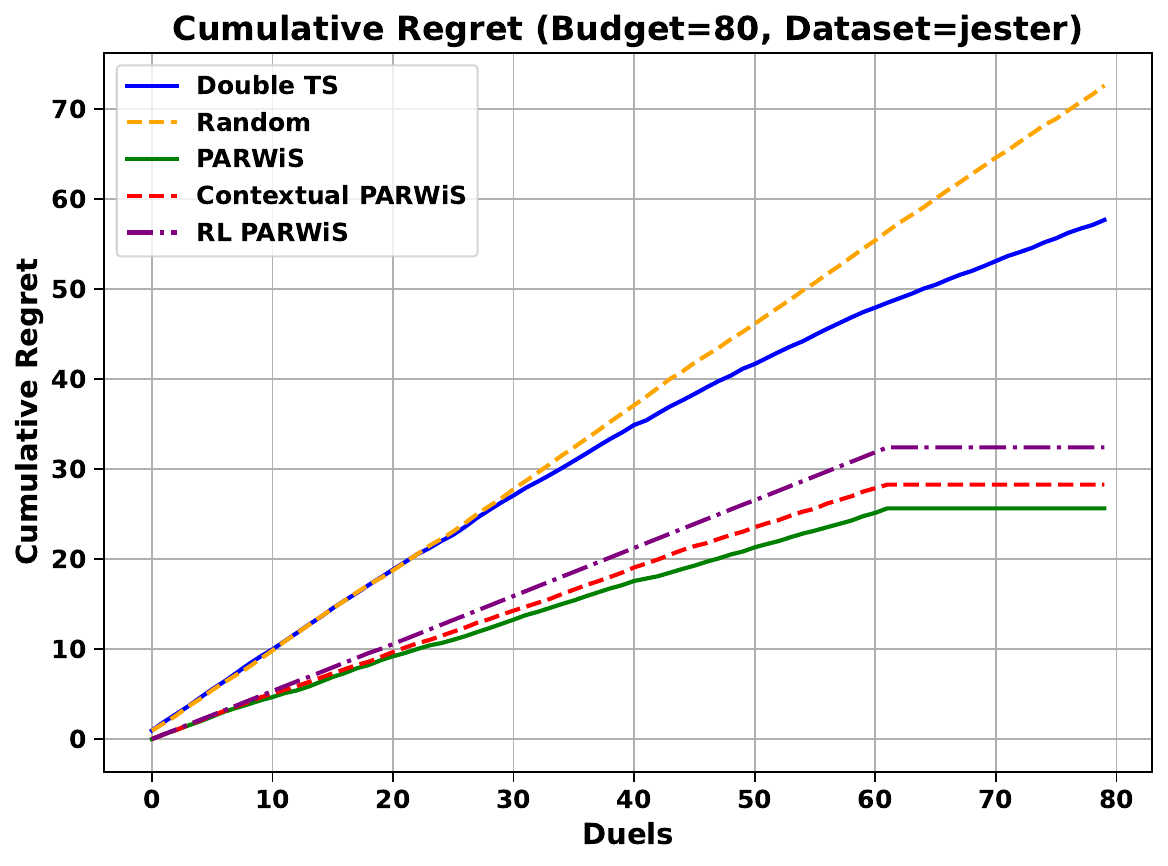}
    \includegraphics[width=0.24\textwidth]{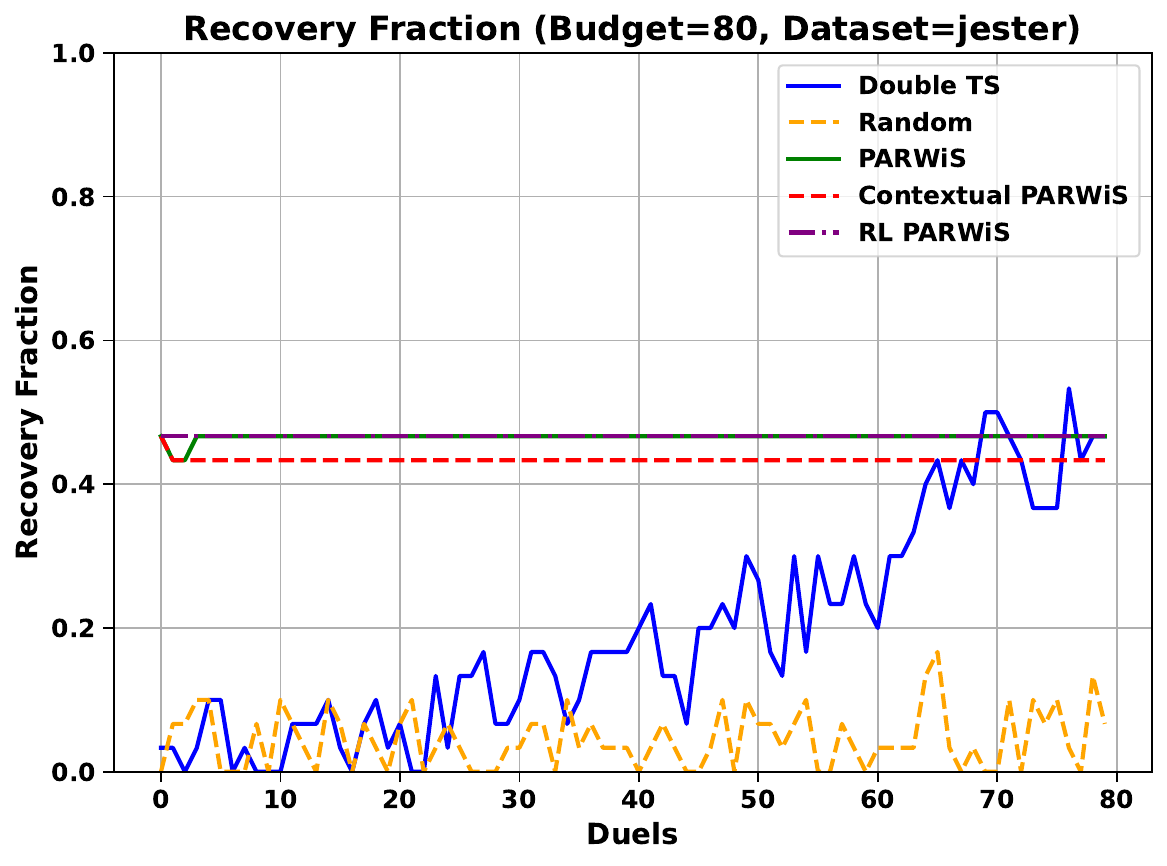}
    \includegraphics[width=0.25\textwidth]{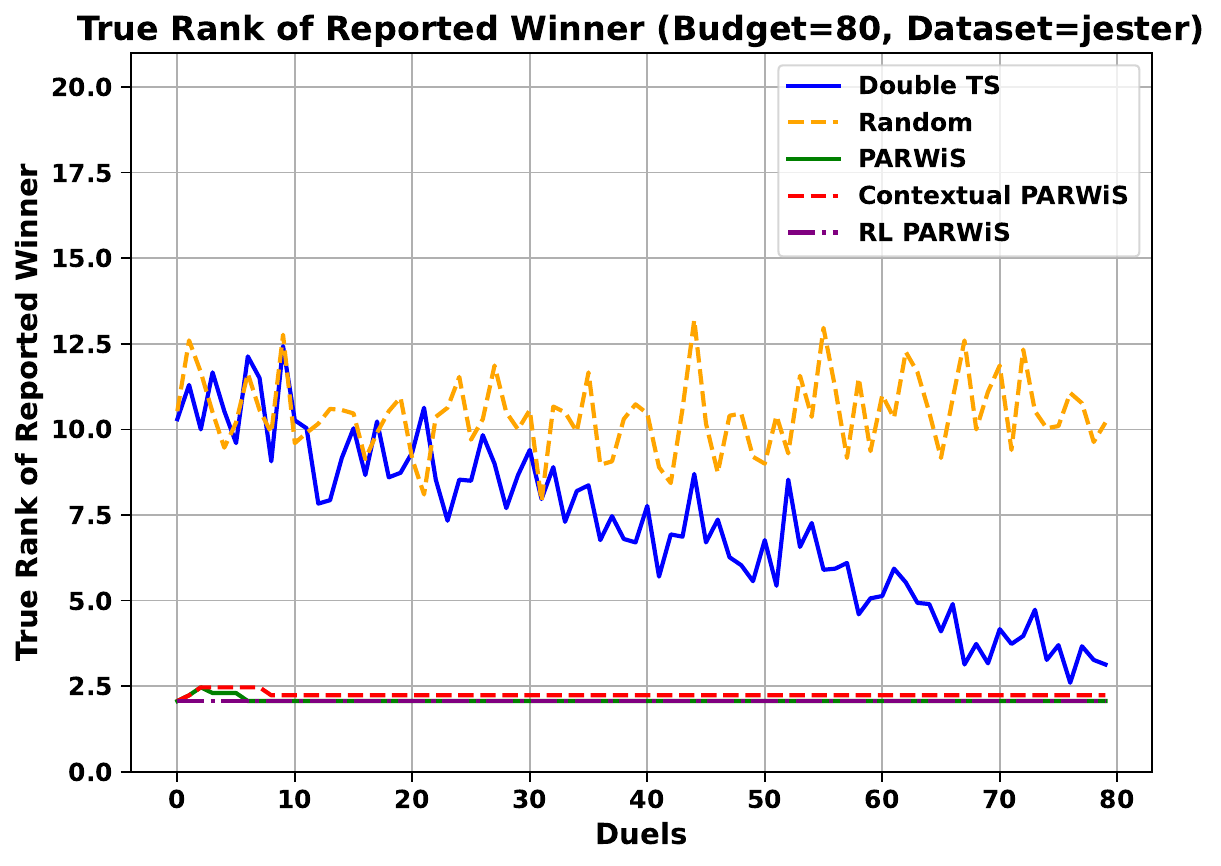}
    \includegraphics[width=0.25\textwidth]{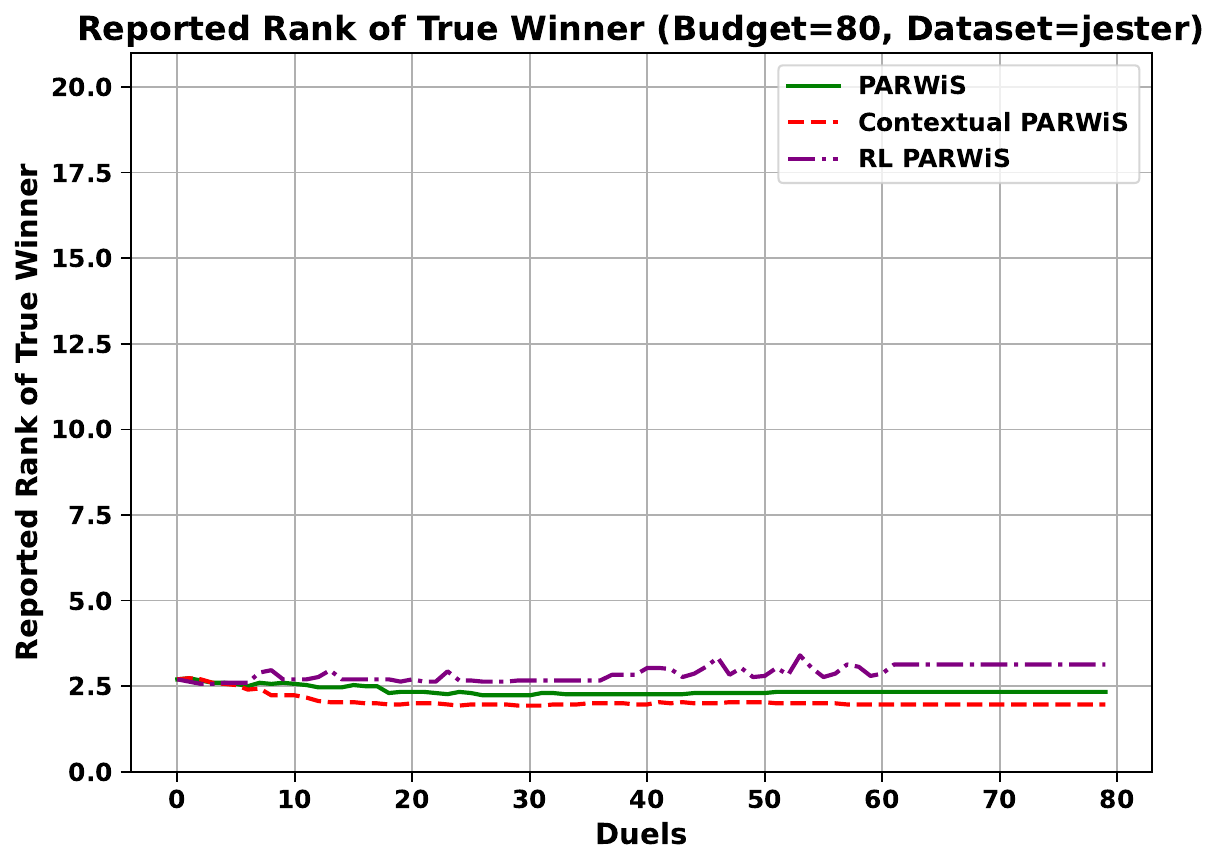}
    \caption{Performance on Jester Dataset at \( B=80 \). From left to right: Cumulative Regret, Recovery Fraction, True Rank of Reported Winner, Reported Rank of True Winner.}
    \label{fig:jester_plots_80}
\end{figure*}

\begin{figure*}[t]
    \centering
    \includegraphics[width=0.24\textwidth]{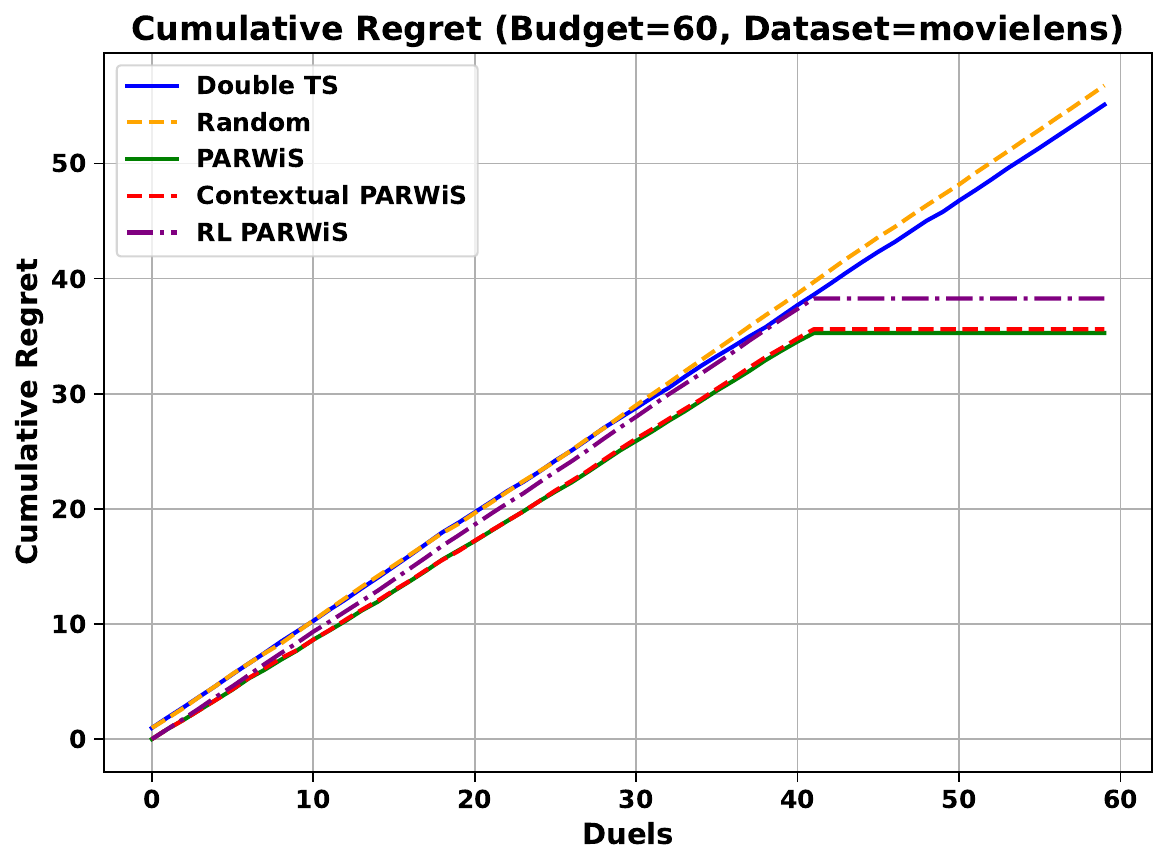}
    \includegraphics[width=0.24\textwidth]{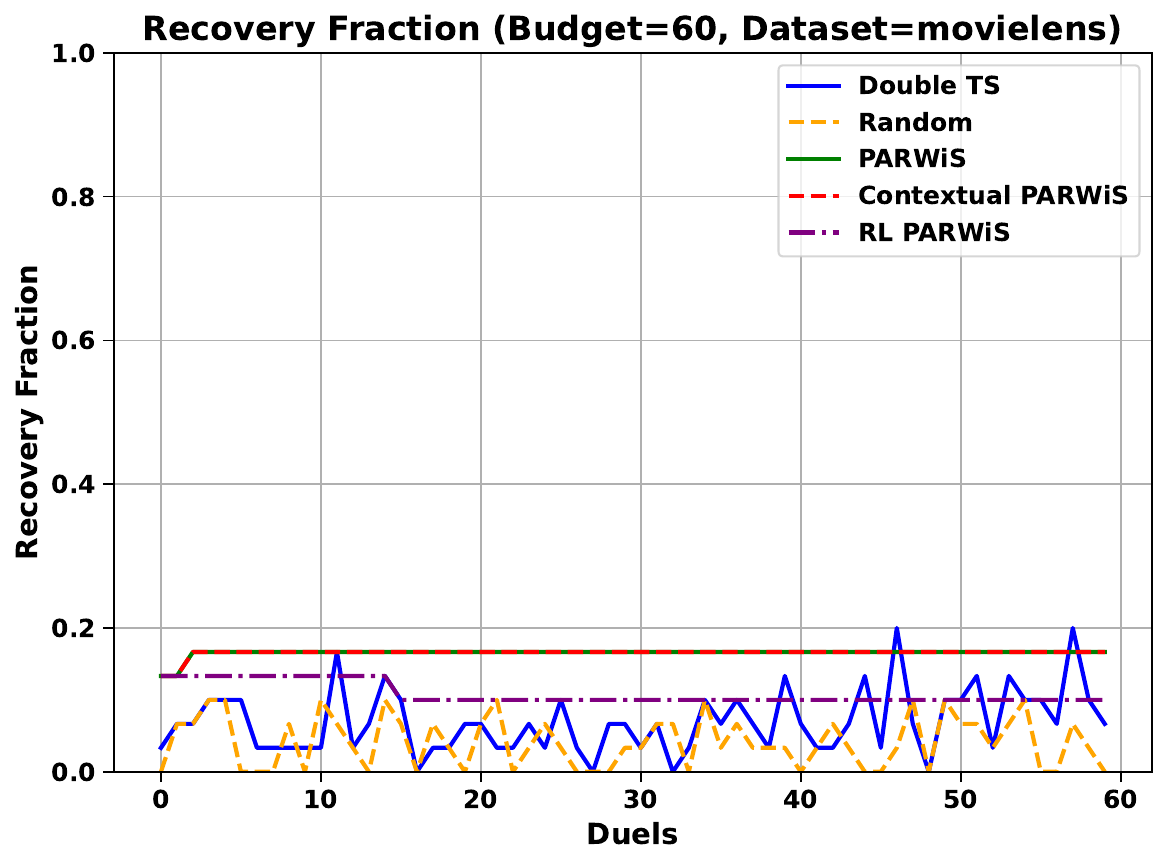}
    \includegraphics[width=0.25\textwidth]{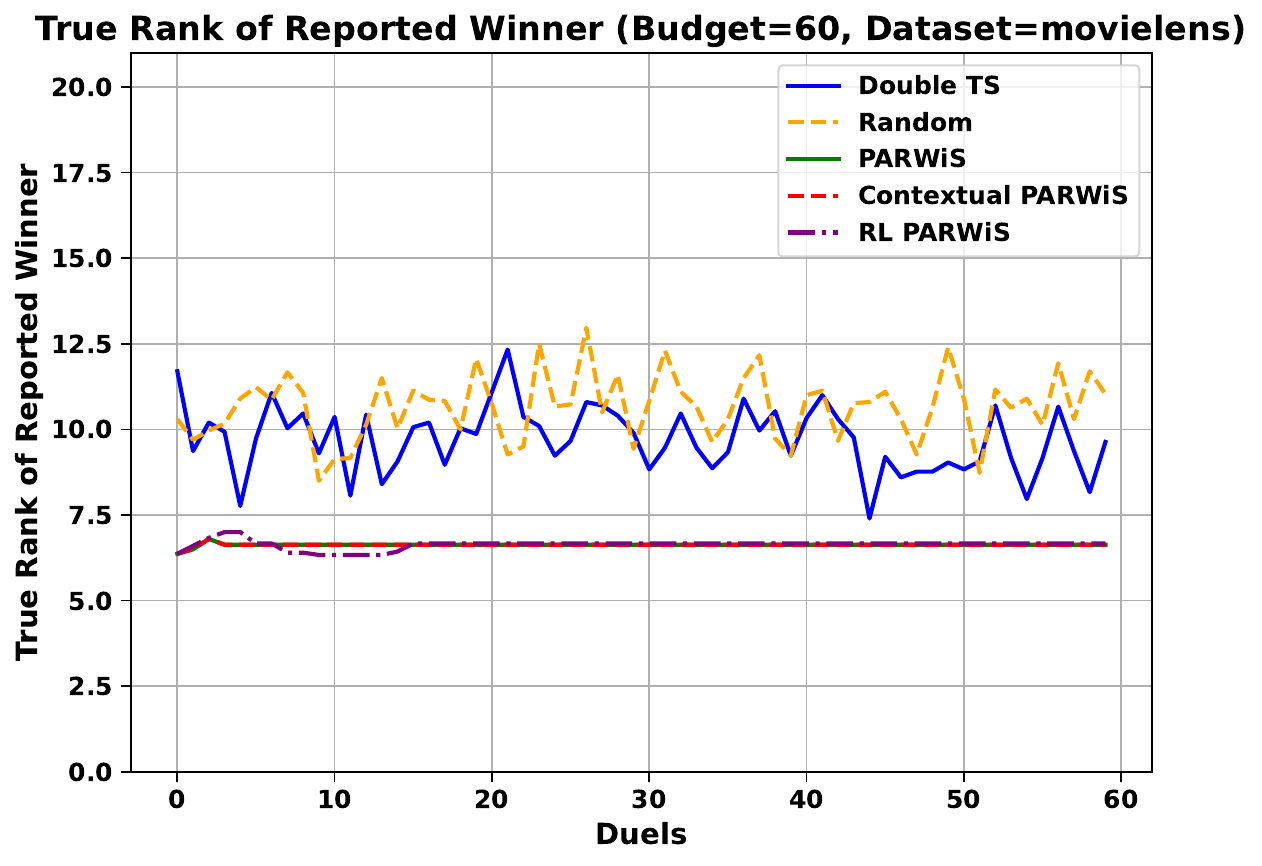}
    \includegraphics[width=0.25\textwidth]{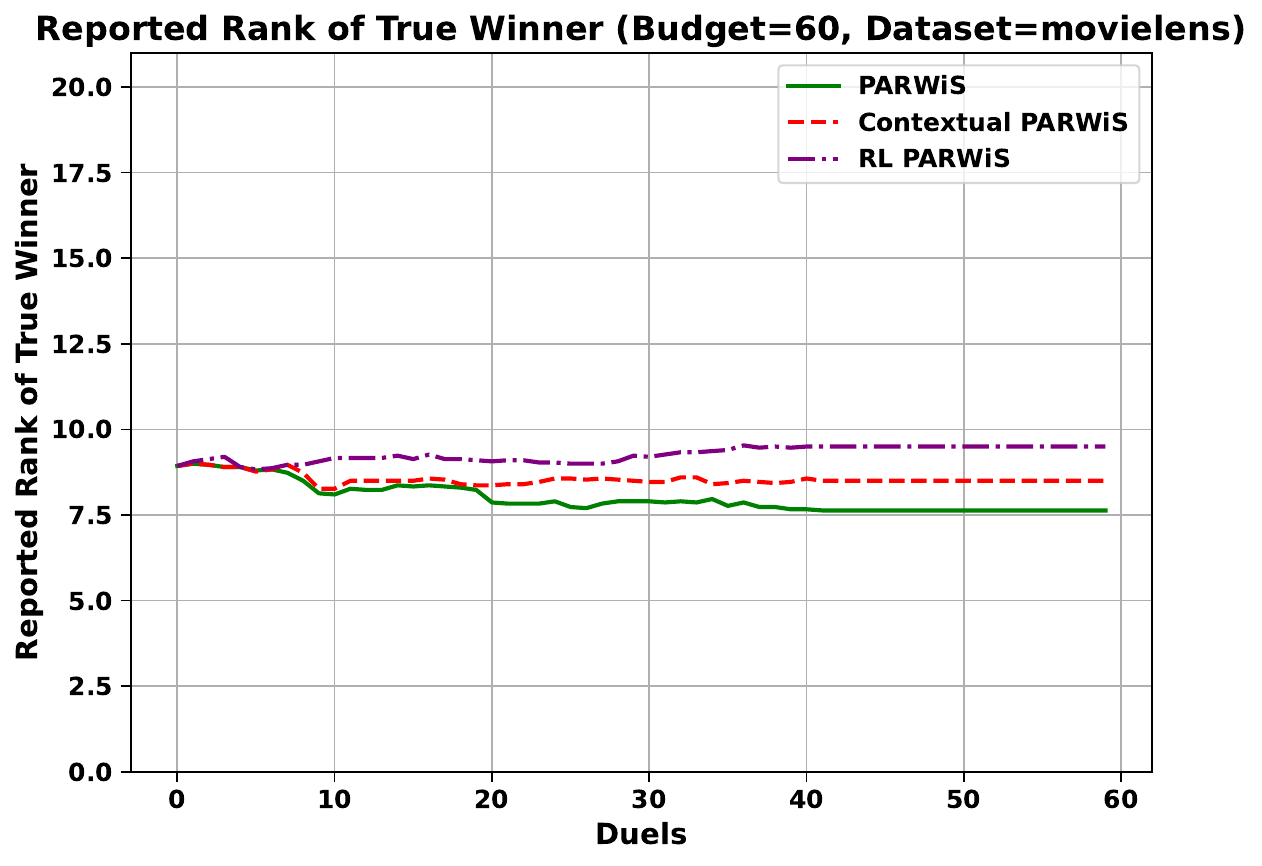}
    \caption{Performance on MovieLens Dataset at \( B=60 \). From left to right: Cumulative Regret, Recovery Fraction, True Rank of Reported Winner, Reported Rank of True Winner.}
    \label{fig:movielens_plots_60}
\end{figure*}

\begin{figure*}[t]
    \centering
    \includegraphics[width=0.24\textwidth]{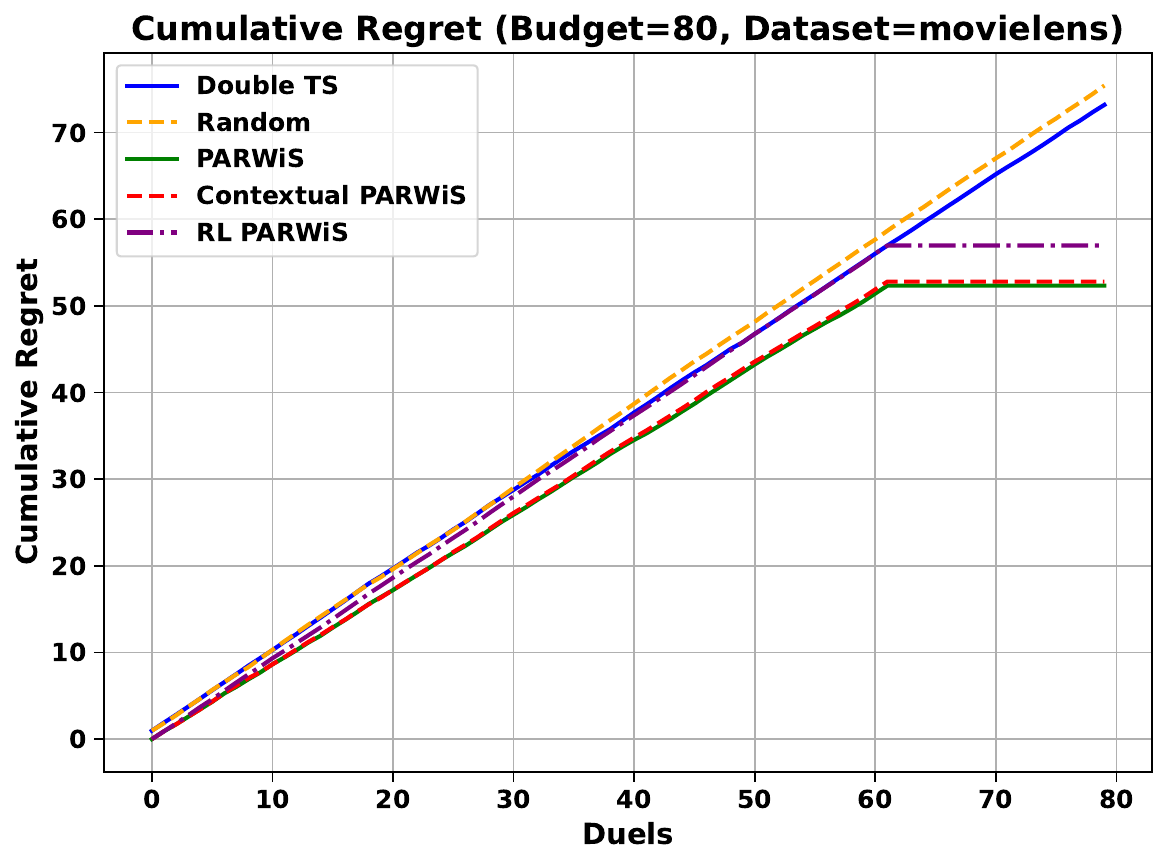}
    \includegraphics[width=0.24\textwidth]{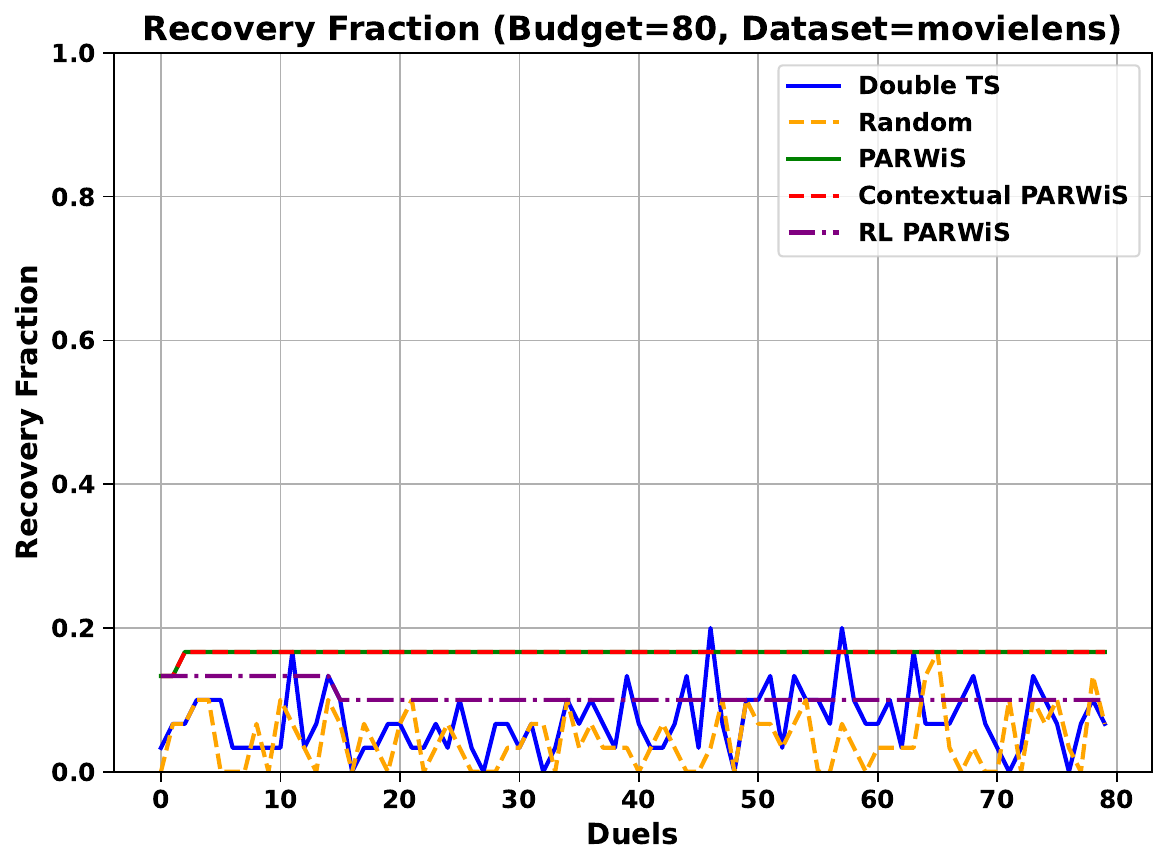}
    \includegraphics[width=0.25\textwidth]{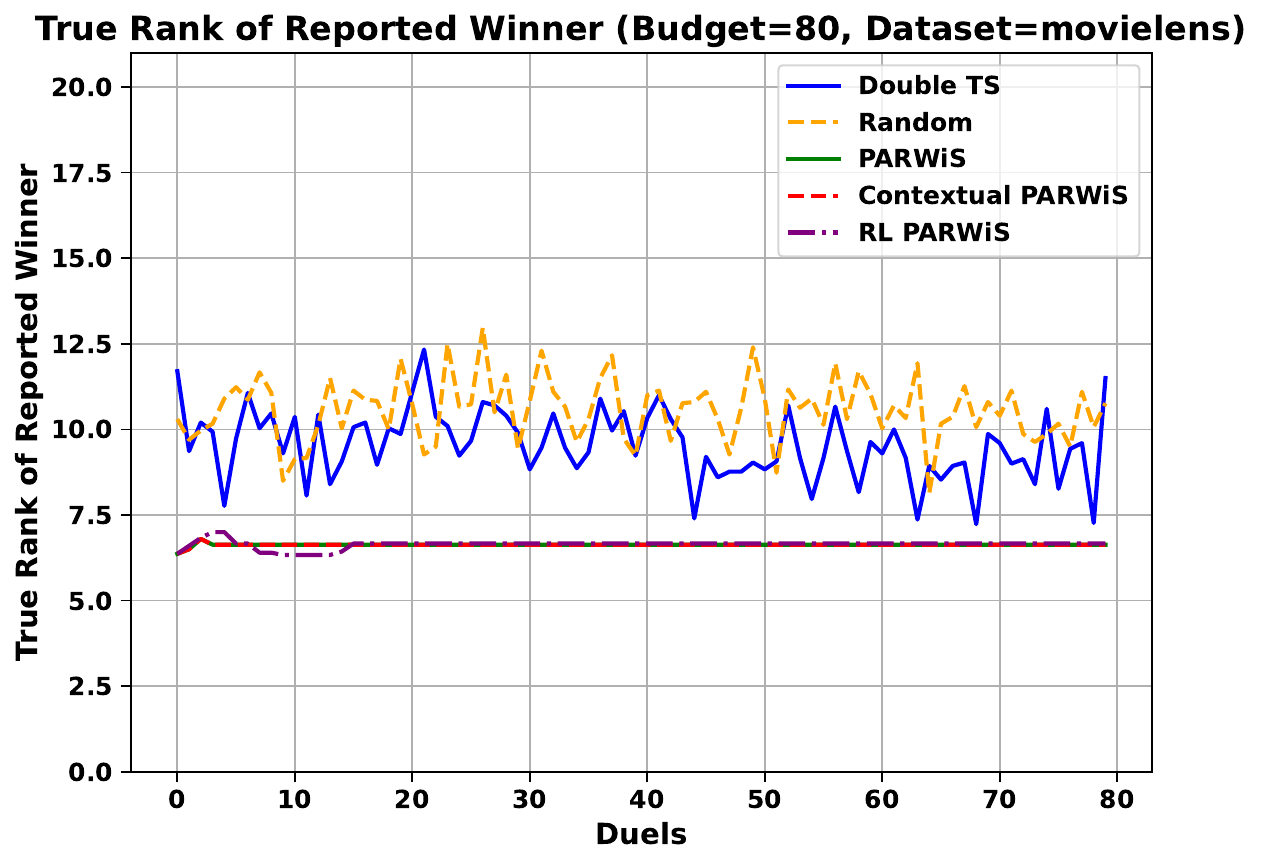}
    \includegraphics[width=0.25\textwidth]{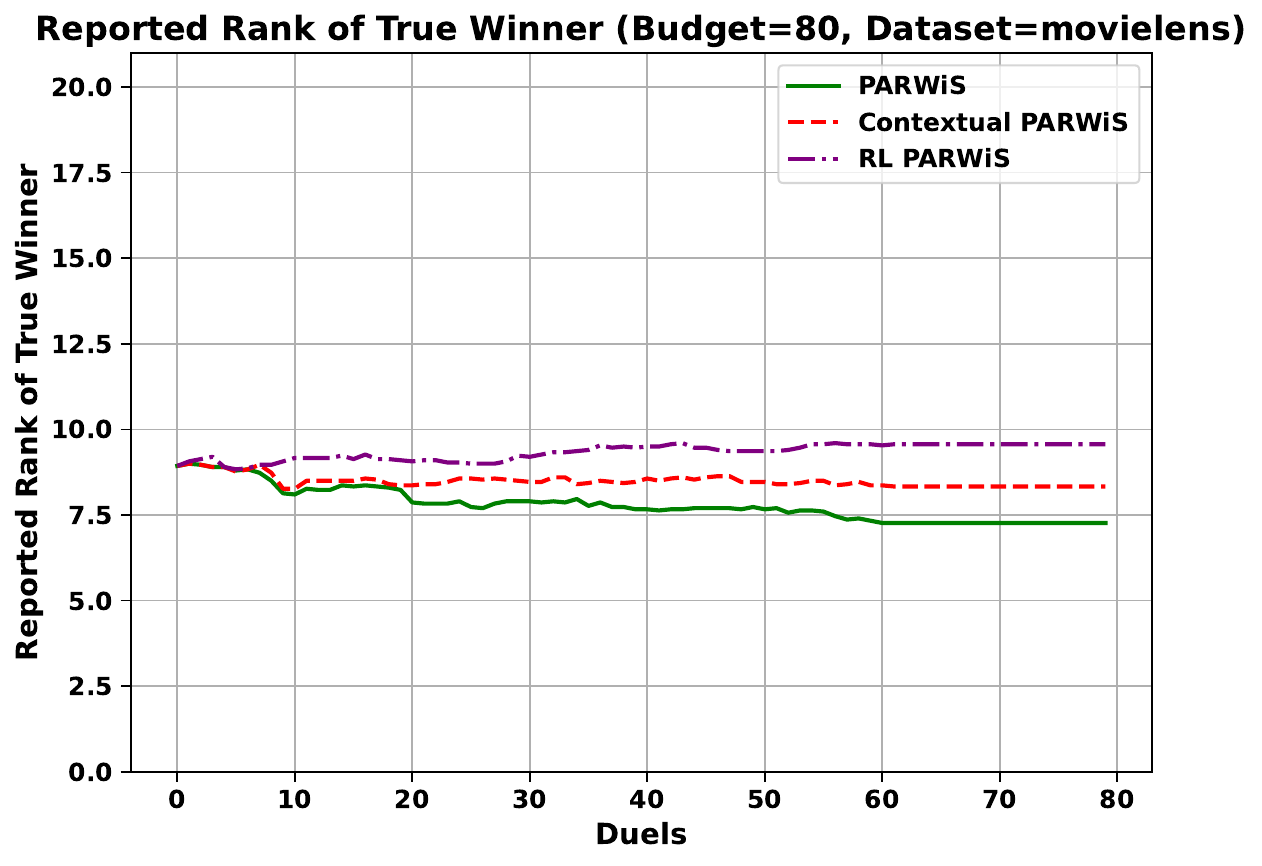}
    \caption{Performance on MovieLens Dataset at \( B=80 \). From left to right: Cumulative Regret, Recovery Fraction, True Rank of Reported Winner, Reported Rank of True Winner.}
    \label{fig:movielens_plots_80}
\end{figure*}

This section provides a comprehensive set of additional visualizations to supplement the performance analysis presented in the main document. Specifically, it includes performance plots and boxplots for the Synthetic, Jester, and MovieLens datasets across budgets \( B=40, 60, 80 \). The performance plots extend the analysis in Figures~\ref{fig:synthetic_plots_40}, \ref{fig:jester_plots_40}, and \ref{fig:movielens_plots_40} by presenting results for \( B=60 \) and \( B=80 \), allowing for a deeper understanding of how the algorithms perform as the budget increases. Each performance plot comprises four subplots: Cumulative Regret, Recovery Fraction, True Rank of Reported Winner, and Reported Rank of True Winner, which collectively illustrate the evolution of these metrics over the number of duels. These plots highlight the dynamic behavior of each algorithm, showing how quickly they converge to the true winner and how regret accumulates over time. The boxplots complement these plots by displaying the distribution of final metric values across the 30 runs, offering a detailed view of variability and consistency in performance at the end of each simulation. Together, these figures provide a holistic view of the algorithms' behavior, capturing both temporal trends and the robustness of their final outcomes across different datasets and budgets.

The performance plots for \( B=60 \) and \( B=80 \) reveal how the algorithms adapt to larger budgets, building on the trends observed at \( B=40 \) in the main document. For the Synthetic dataset (Figures~\ref{fig:synthetic_plots_60} and \ref{fig:synthetic_plots_80}), PARWiS and RL PARWiS continue to outperform the baselines (Double TS and Random) in terms of cumulative regret and recovery fraction, maintaining their lead as the budget increases. The true rank of the reported winner remains low for PARWiS and RL PARWiS, indicating consistent identification of items close to the true winner, while the reported rank of the true winner for RL PARWiS shows slight fluctuations, suggesting potential for further optimization in its ranking mechanism. On the Jester dataset (Figures~\ref{fig:jester_plots_60} and \ref{fig:jester_plots_80}), the larger \(\Delta_{1,2} = 0.0946\) facilitates easier winner identification, and both PARWiS and RL PARWiS achieve stable recovery fractions around 0.467 across all budgets, with minimal regret accumulation after the initial phase. Double TS shows improvement at \( B=80 \), catching up to PARWiS and RL PARWiS in recovery fraction (0.467), as noted in the main text. For the MovieLens dataset (Figures~\ref{fig:movielens_plots_60} and \ref{fig:movielens_plots_80}), the small \(\Delta_{1,2} = 0.0008\) poses a significant challenge, and all algorithms exhibit lower recovery fractions (0.100–0.167). PARWiS maintains the lowest cumulative regret, but the gap between algorithms narrows as the budget increases, reflecting the difficulty of distinguishing the top items in this dataset. These plots collectively demonstrate the robustness of PARWiS and RL PARWiS across varying problem difficulties and budgets, while also highlighting the limitations of all algorithms on challenging datasets like MovieLens.


\begin{figure}[t]
    \centering
    \includegraphics[width=0.45\textwidth]{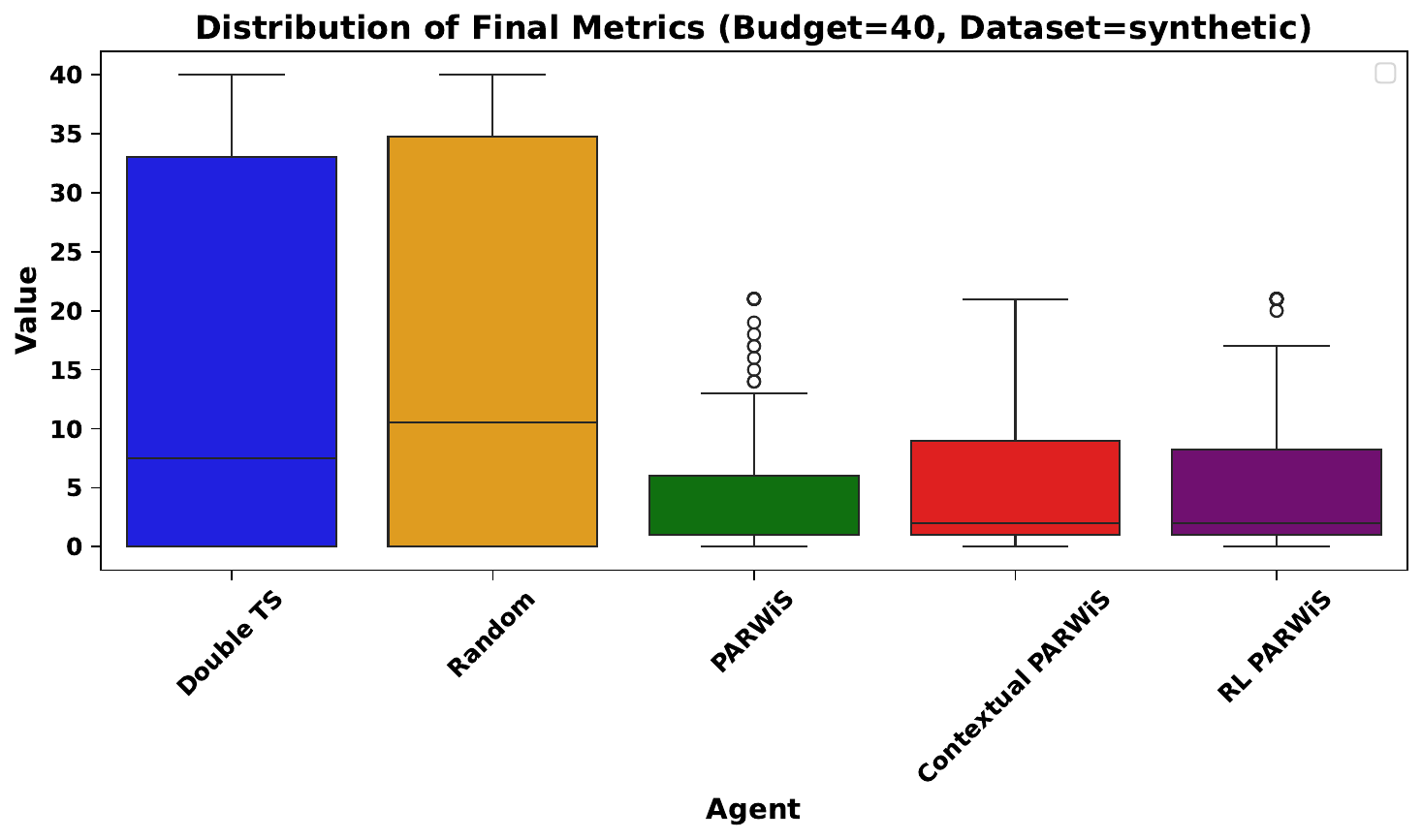}\\
    \includegraphics[width=0.45\textwidth]{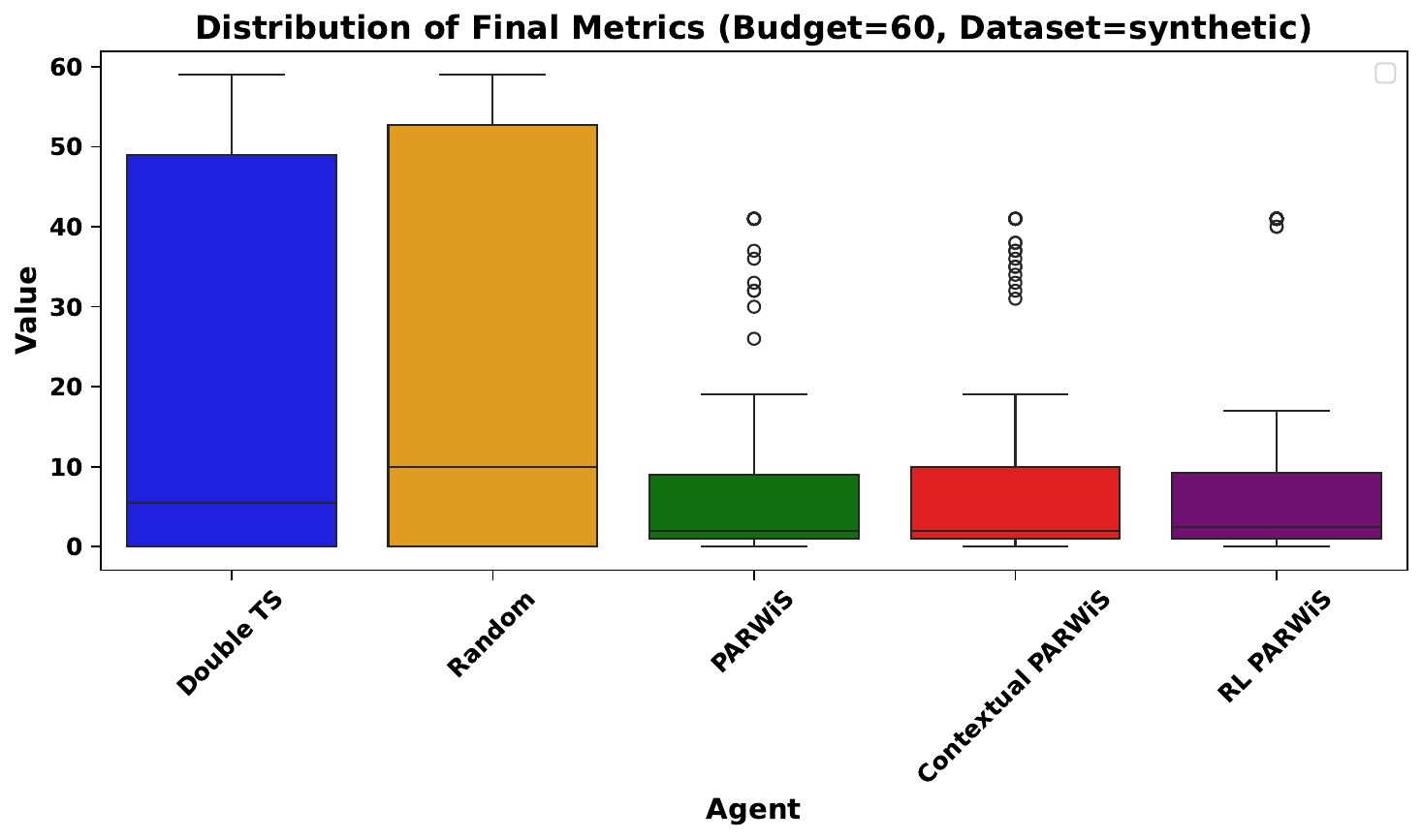}\\
    \includegraphics[width=0.45\textwidth]{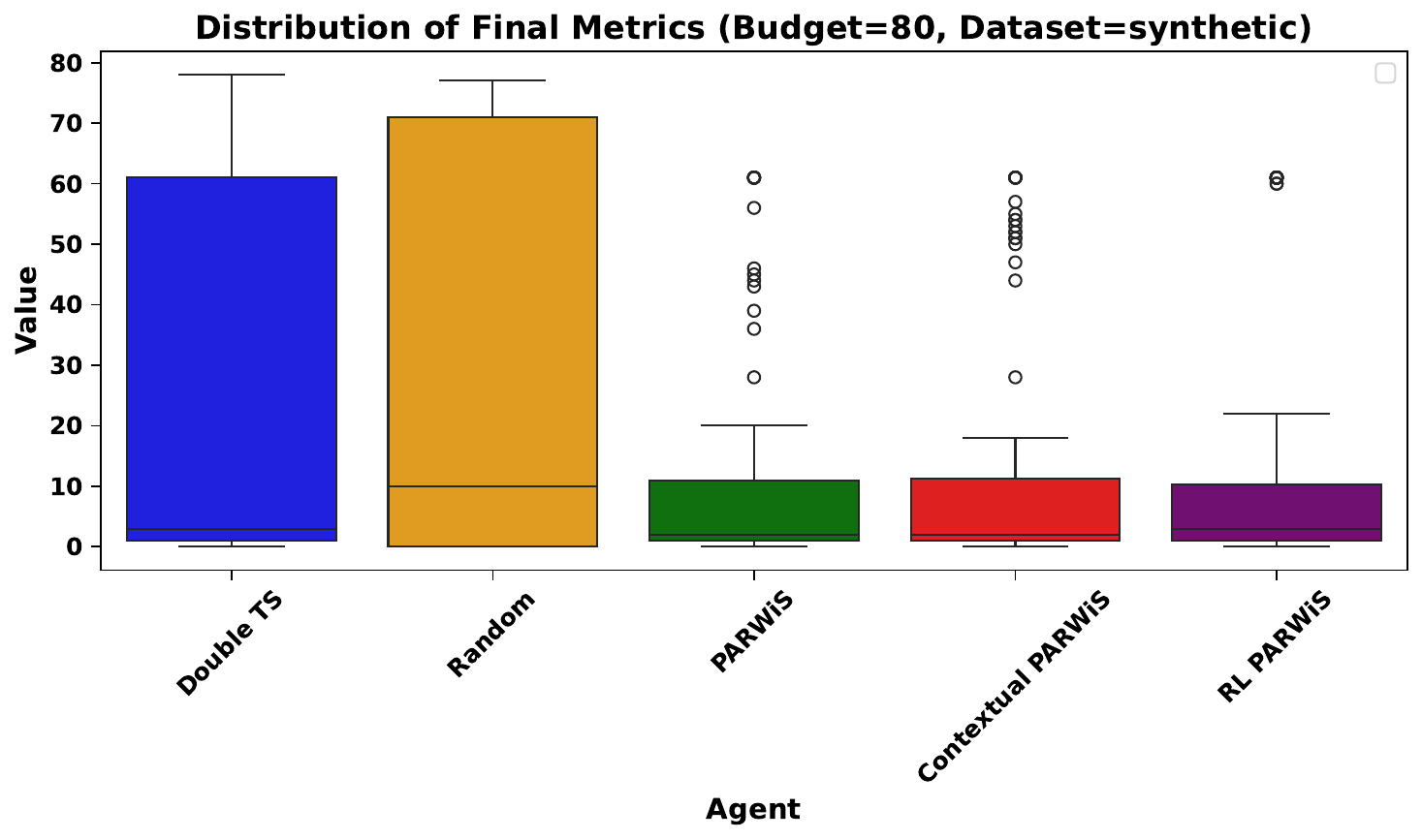}
    \caption{Distribution of final metric values for Synthetic dataset across runs at budgets \( B=40, 60, 80 \). Metrics include Cumulative Regret, Recovery Fraction, True Rank of Reported Winner, and Reported Rank of True Winner.}
    \label{fig:synthetic_boxplots}
\end{figure}

\begin{figure}[t]
    \centering
    \includegraphics[width=0.495\textwidth]{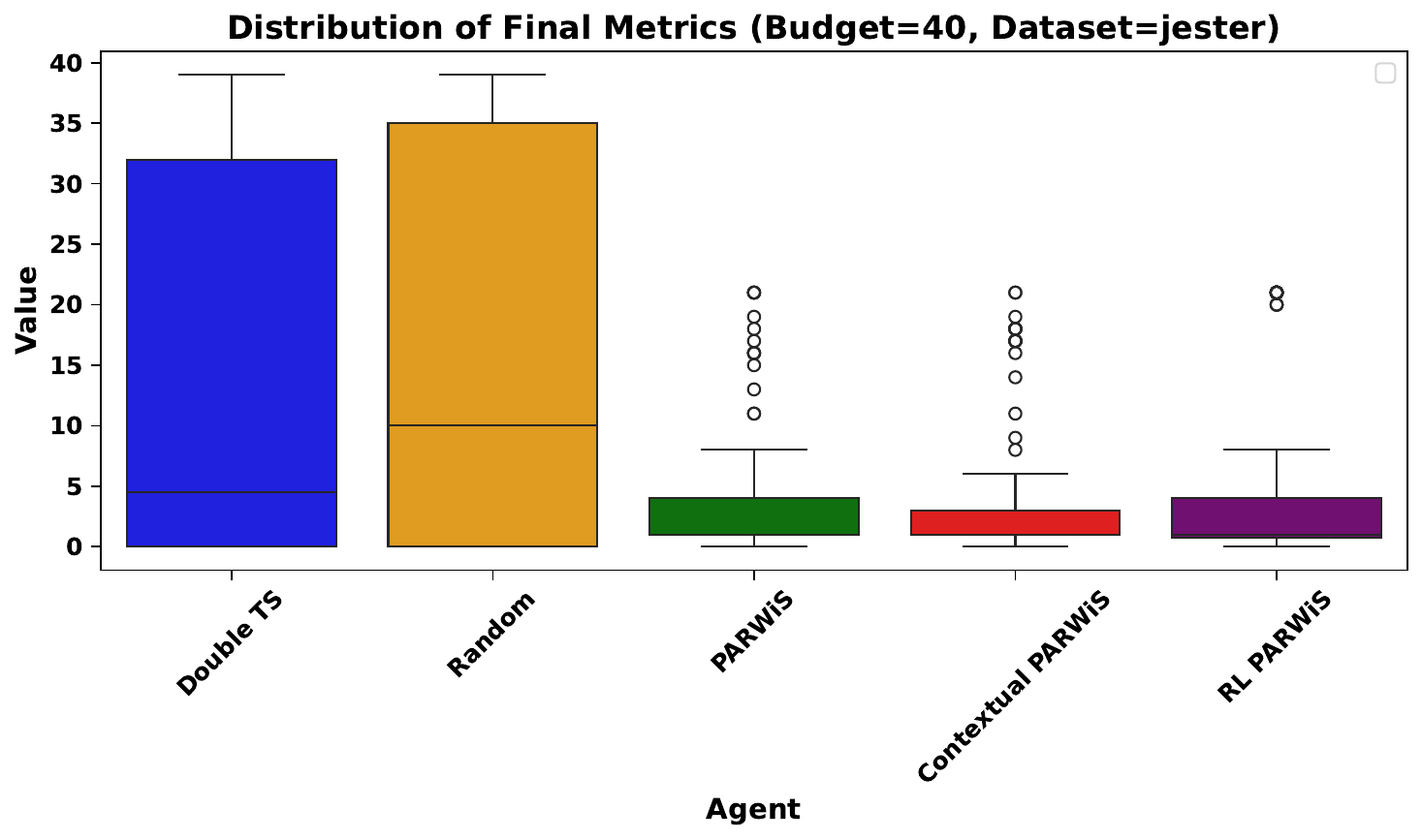}
    \includegraphics[width=0.495\textwidth]{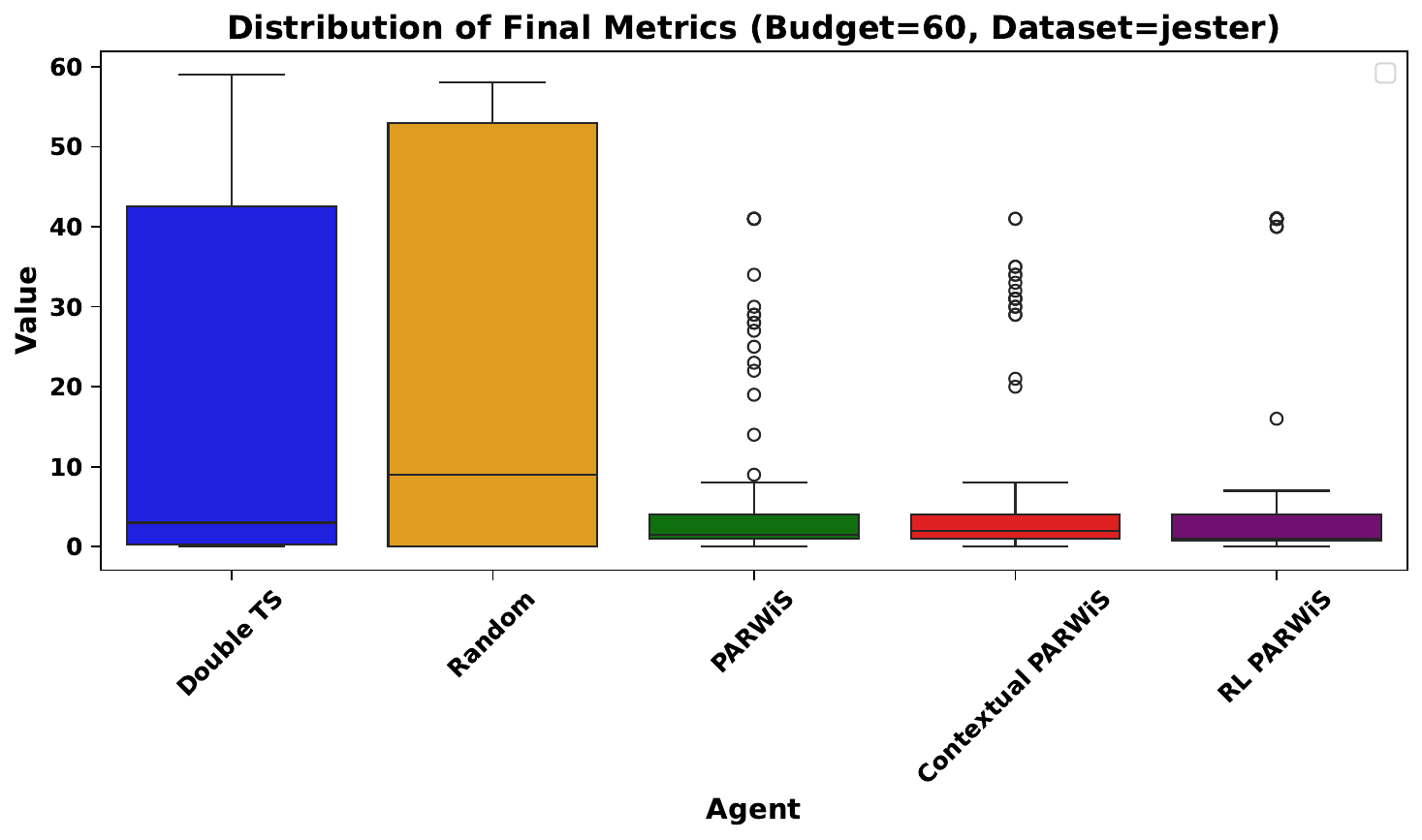}
    \includegraphics[width=0.495\textwidth]{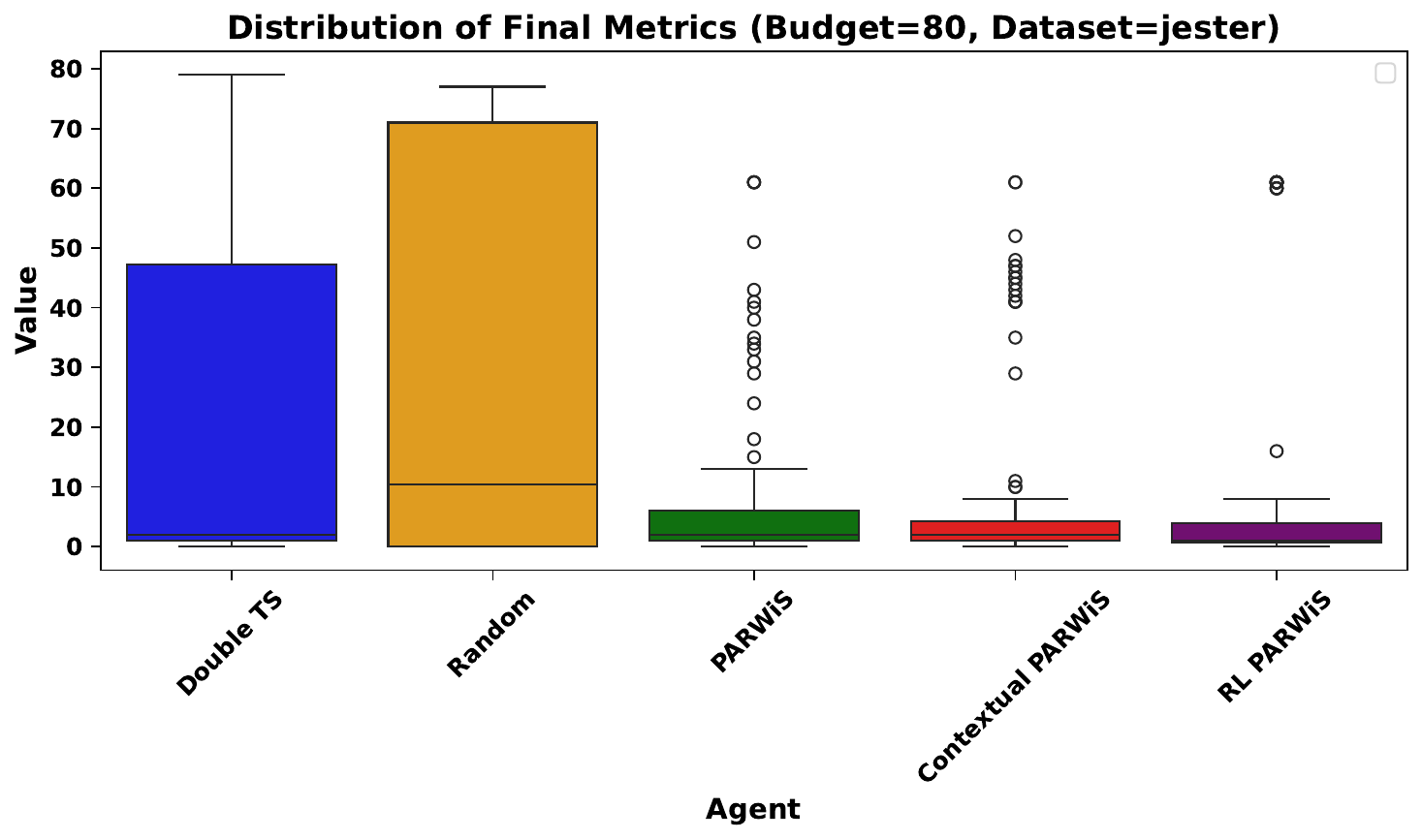}
    \caption{Distribution of final metric values for Jester dataset across runs at budgets \( B=40, 60, 80 \). Metrics include Cumulative Regret, Recovery Fraction, True Rank of Reported Winner, and Reported Rank of True Winner.}
    \label{fig:jester_boxplots}
\end{figure}

\begin{figure}[t]
    \centering
    \includegraphics[width=0.495\textwidth]{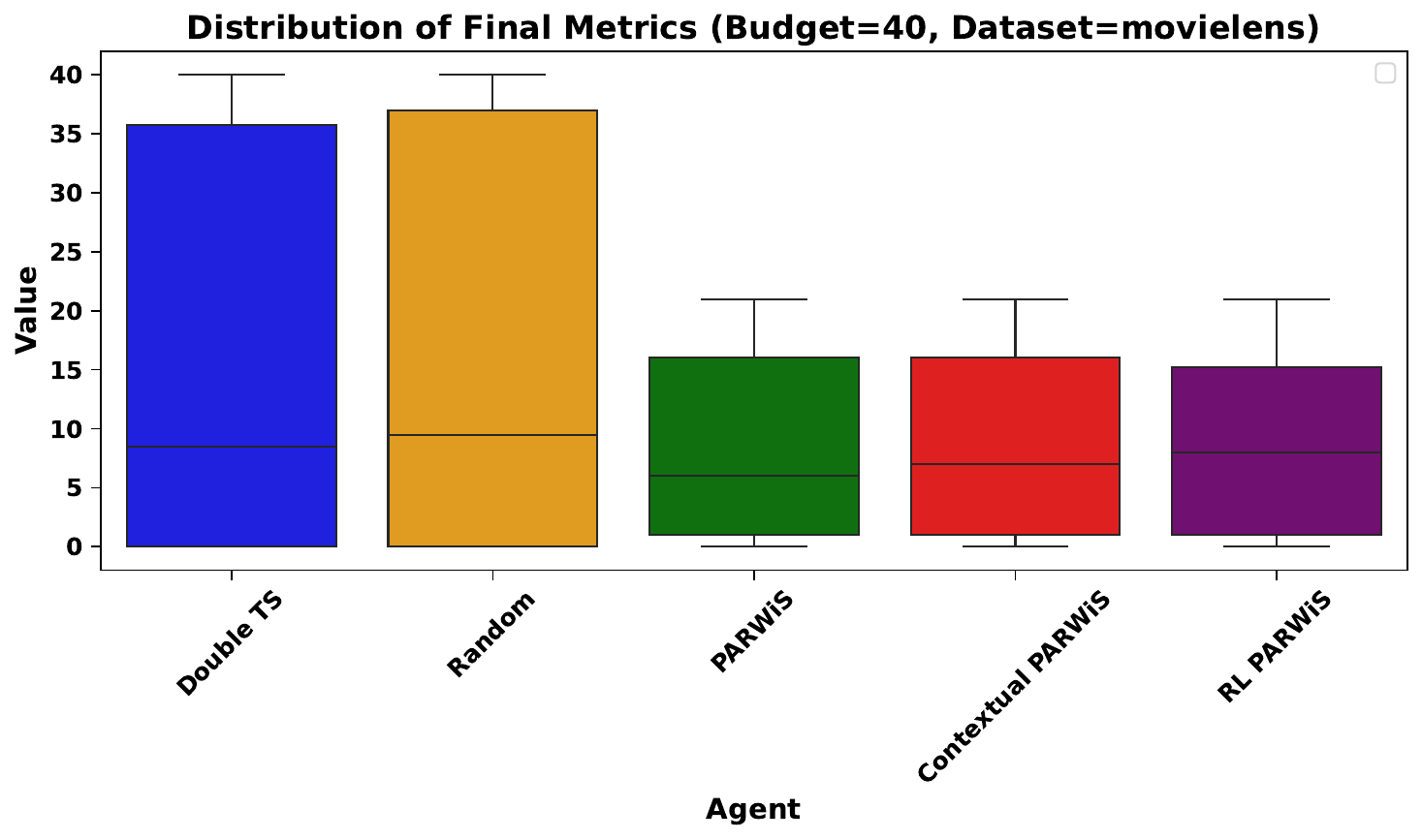}
    \includegraphics[width=0.495\textwidth]{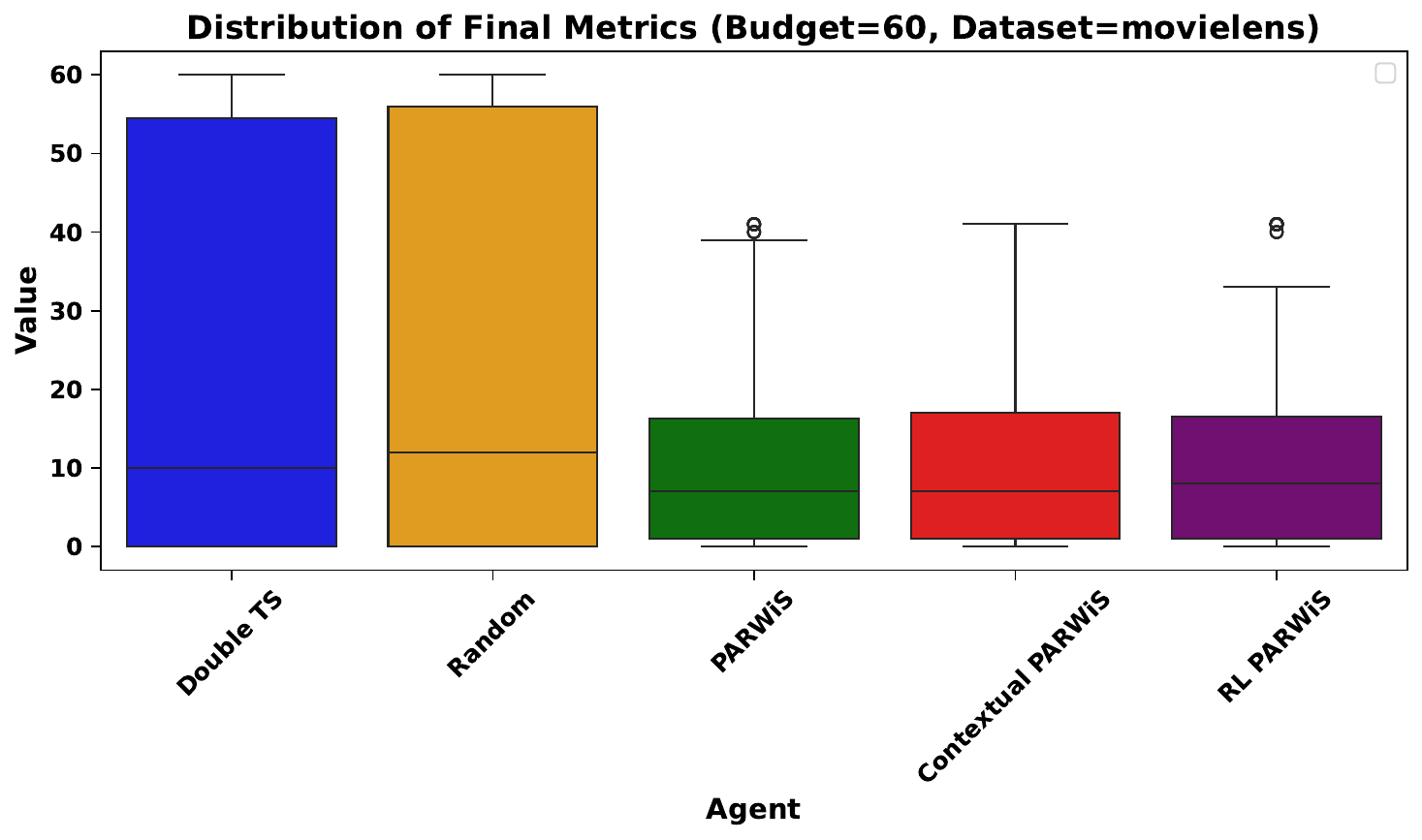}
    \includegraphics[width=0.495\textwidth]{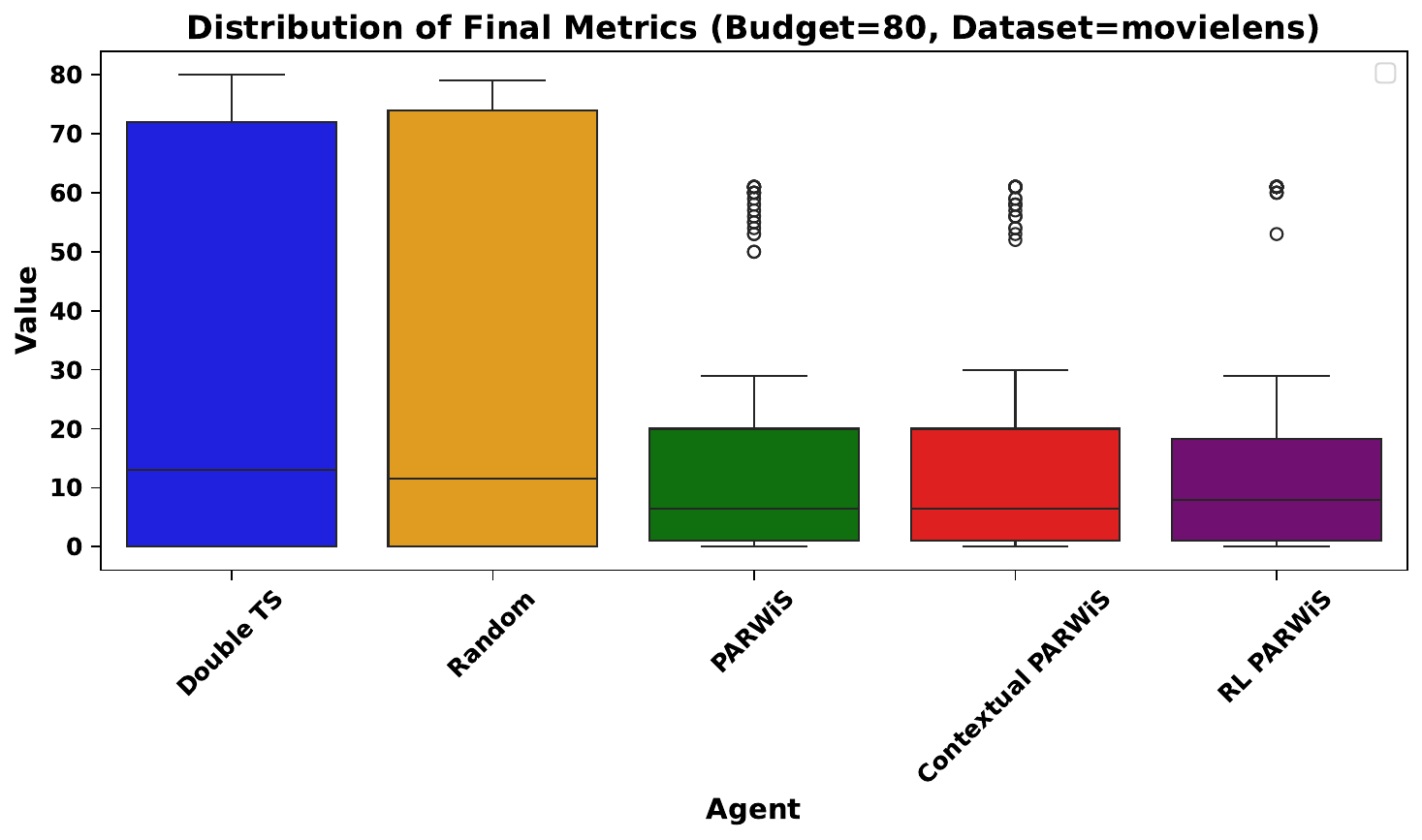}
    \caption{Distribution of final metric values for MovieLens dataset across runs at budgets \( B=40, 60, 80 \). Metrics include Cumulative Regret, Recovery Fraction, True Rank of Reported Winner, and Reported Rank of True Winner.}
    \label{fig:movielens_boxplots}
\end{figure}

\subsection{Distribution of final metrics}

The boxplots in Figures~\ref{fig:synthetic_boxplots}, \ref{fig:jester_boxplots}, and \ref{fig:movielens_boxplots} illustrate the distribution of final metric values (Cumulative Regret, Recovery Fraction, True Rank of Reported Winner, and Reported Rank of True Winner) across the 30 runs for each dataset and budget. These plots provide a detailed view of the variability in performance at the end of each simulation, complementing the mean trends shown in the performance plots. For the Synthetic dataset (Figure~\ref{fig:synthetic_boxplots}), PARWiS and RL PARWiS exhibit tighter distributions in recovery fraction and true rank, indicating more consistent performance compared to Double TS and Random, which show wider spreads and more outliers, particularly in cumulative regret. On the Jester dataset (Figure~\ref{fig:jester_boxplots}), the distributions for PARWiS and RL PARWiS are notably compact, with recovery fractions centered around 0.467 and minimal variability in true rank (around 2.067), reflecting the easier problem (\(\Delta_{1,2} = 0.0946\)). Double TS shows increased consistency at \( B=80 \), aligning with its improved recovery fraction. For the MovieLens dataset (Figure~\ref{fig:movielens_boxplots}), the distributions are broader for all algorithms due to the challenging \(\Delta_{1,2} = 0.0008\), with PARWiS and Contextual PARWiS showing slightly less variability in true rank (around 6.633) compared to RL PARWiS (6.667). These boxplots highlight the robustness of PARWiS and RL PARWiS across runs, while also underscoring the impact of problem difficulty on performance consistency, particularly on MovieLens where all algorithms exhibit higher variability.

\begin{table*}[t]
\centering
\caption{Error analysis across synthetic and Jester datasets.}
\label{tab:error_analysis_synthetic_jester}
\begin{tabular}{lcccccccccccc}
\toprule
& \multicolumn{6}{c}{Synthetic (\(\Delta_{1,2} = 0.0152 \pm 0.0190\))} & \multicolumn{6}{c}{Jester (\(\Delta_{1,2} = 0.0946 \pm 0.0000\))} \\
\cmidrule(lr){2-7} \cmidrule(lr){8-13}
& \multicolumn{2}{c}{\( B=40 \)} & \multicolumn{2}{c}{\( B=60 \)} & \multicolumn{2}{c}{\( B=80 \)} & \multicolumn{2}{c}{\( B=40 \)} & \multicolumn{2}{c}{\( B=60 \)} & \multicolumn{2}{c}{\( B=80 \)} \\
\cmidrule(lr){2-3} \cmidrule(lr){4-5} \cmidrule(lr){6-7} \cmidrule(lr){8-9} \cmidrule(lr){10-11} \cmidrule(lr){12-13}
Agent & Fail & Avg TR & Fail & Avg TR & Fail & Avg TR & Fail & Avg TR & Fail & Avg TR & Fail & Avg TR \\
\midrule
Double TS & 0.800 & 10.042 & 0.933 & 6.929 & 0.733 & 6.136 & 0.833 & 7.840 & 0.767 & 6.304 & 0.533 & 5.250 \\
Random & 0.967 & 11.103 & 0.933 & 10.357 & 1.000 & 10.733 & 0.967 & 11.069 & 1.000 & 9.367 & 0.933 & 11.071 \\
PARWiS & \textbf{0.533} & 5.188 & \textbf{0.533} & 5.188 & \textbf{0.533} & 5.188 & \textbf{0.533} & \textbf{3.000} & \textbf{0.533} & \textbf{3.000} & \textbf{0.533} & \textbf{3.000} \\
Contextual PARWiS & 0.633 & 5.842 & 0.633 & 5.842 & 0.633 & 5.842 & 0.567 & 3.176 & 0.567 & 3.176 & 0.567 & 3.176 \\
RL PARWiS & 0.633 & \textbf{5.000} & 0.633 & \textbf{5.000} & 0.633 & \textbf{5.000} & \textbf{0.533} & \textbf{3.000} & \textbf{0.533} & \textbf{3.000} & \textbf{0.533} & \textbf{3.000} \\
\bottomrule
\end{tabular}
\end{table*}

\subsection{Additional tables}
\label{app:tables}

This section presents additional tables that provide deeper insights into the performance of the algorithms, complementing the results and discussions in the main document. The tables include the reported rank of the true winner, statistical significance tests, and error analysis across the Synthetic, Jester, and MovieLens datasets for budgets \( B=40, 60, 80 \). These tables offer a quantitative perspective on the algorithms' ranking accuracy, the significance of performance differences, and the consistency of their winner identification, particularly in failure cases. Together, they enhance the understanding of the algorithms' behavior under varying problem difficulties and budget constraints, providing a more comprehensive evaluation of their effectiveness and robustness.

Table~\ref{tab:reportedrank_comparison} reports the average reported rank of the true winner, as determined by the internal rankings of PARWiS, Contextual PARWiS, and RL PARWiS, across all datasets and budgets. This metric, applicable only to these algorithms due to their use of spectral ranking, indicates how well the true winner is ranked within the agent’s estimated ordering (lower is better). For the Synthetic dataset (\(\Delta_{1,2} = 0.0152 \pm 0.0190\)), PARWiS consistently achieves the lowest reported rank (e.g., 4.400 at \( B=40 \), decreasing to 4.267 at \( B=60, 80 \)), reflecting its accurate estimation of the true winner’s position. Contextual PARWiS performs similarly (4.433 at \( B=40 \), 4.233 at \( B=60, 80 \)), while RL PARWiS has a higher reported rank (e.g., 5.333 at \( B=40 \)), indicating potential for improvement in its ranking mechanism, as noted in the main text. On the Jester dataset (\(\Delta_{1,2} = 0.0946\)), the easier problem allows all algorithms to rank the true winner higher, with Contextual PARWiS achieving the best performance (e.g., 2.000 at \( B=40 \), 1.967 at \( B=60 \)). For the MovieLens dataset (\(\Delta_{1,2} = 0.0008\)), the reported ranks are higher due to the challenging separation, with PARWiS performing best (e.g., 7.833 at \( B=40 \), improving to 7.267 at \( B=80 \)), while RL PARWiS struggles (e.g., 9.567 at \( B=80 \)). These results highlight PARWiS’s superior ranking accuracy across datasets, particularly on MovieLens, where the small \(\Delta_{1,2}\) makes winner identification difficult.

Table~\ref{tab:ttest_recovery_comparison} presents the results of pairwise t-tests comparing the recovery fraction of Double TS and PARWiS across all datasets and budgets, assessing the statistical significance of their performance differences. The recovery fraction, defined as the fraction of runs where the true winner is recommended, is a key metric of success in this study. On the Synthetic dataset, PARWiS significantly outperforms Double TS at \( B=40 \) (t-stat = -2.246, p-value = 0.029) and \( B=60 \) (t-stat = -3.862, p-value = 0.000), with the gap narrowing at \( B=80 \) (t-stat = -1.616, p-value = 0.112) as Double TS improves with more comparisons. For the Jester dataset, the difference is significant at \( B=40 \) (t-stat = -2.594, p-value = 0.012) and marginally significant at \( B=60 \) (t-stat = -1.921, p-value = 0.060), but diminishes at \( B=80 \) (t-stat = 0.000, p-value = 1.000) where Double TS catches up with a recovery fraction of 0.467, matching PARWiS. On MovieLens, the differences are not significant (e.g., p-value = 0.723 at \( B=40 \), 0.235 at \( B=60, 80 \)), reflecting the dataset’s difficulty (\(\Delta_{1,2} = 0.0008\)), which impacts all algorithms similarly. These tests confirm that PARWiS’s improvements over Double TS are statistically significant in most cases, particularly on Synthetic and Jester datasets with moderate to large \(\Delta_{1,2}\), but less so on MovieLens where the small separation reduces performance differences.

Tables~\ref{tab:error_analysis_synthetic_jester} and \ref{tab:error_analysis_movielens} provide an error analysis by reporting the failure rate (fraction of runs where the true winner is not recommended) and the average true rank of the reported winner in failure cases (Avg TR) for each algorithm across datasets and budgets. This analysis offers insight into the algorithms’ consistency and the severity of their errors when they fail to identify the true winner. For the Synthetic dataset (Table~\ref{tab:error_analysis_synthetic_jester}), PARWiS has the lowest failure rate (0.533 across all budgets), and when it fails, the reported winner’s true rank is relatively low (5.188), indicating that it selects items close to the true winner. RL PARWiS has a slightly higher failure rate (0.633) but fails closer to the true winner (5.000), while Double TS and Random fail more frequently (e.g., 0.800 and 0.967 at \( B=40 \)) with higher true ranks (10.042 and 11.103). On the Jester dataset, the easier problem (\(\Delta_{1,2} = 0.0946\)) results in lower failure rates for PARWiS and RL PARWiS (0.533), with failures very close to the true winner (3.000). Double TS improves at \( B=80 \) (failure rate 0.533, true rank 5.250), aligning with its recovery fraction improvement. For the MovieLens dataset (Table~\ref{tab:error_analysis_movielens}), the small \(\Delta_{1,2} = 0.0008\) leads to high failure rates for all algorithms (0.833–0.967 at \( B=40 \)), but PARWiS and Contextual PARWiS fail closer to the true winner (7.760) compared to RL PARWiS (7.296) and Double TS (9.920 at \( B=40 \), increasing to 12.250 at \( B=80 \)). These tables highlight the robustness of PARWiS and RL PARWiS, as they fail less frequently and closer to the true winner, particularly on easier datasets like Jester, while also showing the challenges posed by MovieLens across all algorithms.

\begin{table*}[h]
\centering
\caption{Reported rank of true winner across datasets and budgets.}
\label{tab:reportedrank_comparison}
\begin{tabular}{lccccccccc}
\toprule
& \multicolumn{3}{c}{Synthetic (\(\Delta_{1,2} = 0.0152 \pm 0.0190\))} & \multicolumn{3}{c}{Jester (\(\Delta_{1,2} = 0.0946 \pm 0.0000\))} & \multicolumn{3}{c}{MovieLens (\(\Delta_{1,2} = 0.0008 \pm 0.0000\))} \\
\cmidrule(lr){2-4} \cmidrule(lr){5-7} \cmidrule(lr){8-10}
Agent & \( B=40 \) & \( B=60 \) & \( B=80 \) & \( B=40 \) & \( B=60 \) & \( B=80 \) & \( B=40 \) & \( B=60 \) & \( B=80 \) \\
\midrule
PARWiS & \textbf{4.400} & \textbf{4.267} & \textbf{4.267} & 2.333 & 2.333 & 2.333 & \textbf{7.833} & \textbf{7.833} & \textbf{7.267} \\
Contextual PARWiS & 4.433 & 4.233 & 4.233 & \textbf{2.000} & \textbf{1.967} & \textbf{2.000} & 8.400 & 8.333 & 8.333 \\
RL PARWiS & 5.333 & 5.433 & 5.433 & 2.633 & 3.133 & 3.133 & 9.100 & 9.100 & 9.567 \\
\bottomrule
\end{tabular}
\end{table*}

\begin{table*}[h]
\centering
\caption{Error analysis across the MovieLens dataset.}
\label{tab:error_analysis_movielens}
\begin{tabular}{lcccccc}
\toprule
& \multicolumn{6}{c}{MovieLens (\(\Delta_{1,2} = 0.0008 \pm 0.0000\))} \\
\cmidrule(lr){2-7}
& \multicolumn{2}{c}{\( B=40 \)} & \multicolumn{2}{c}{\( B=60 \)} & \multicolumn{2}{c}{\( B=80 \)} \\
\cmidrule(lr){2-3} \cmidrule(lr){4-5} \cmidrule(lr){6-7}
Agent & Fail & Avg TR & Fail & Avg TR & Fail & Avg TR \\
\midrule
Double TS & 0.833 & 9.920 & 0.933 & 11.036 & 0.933 & 12.250 \\
Random & 0.967 & 9.552 & 1.000 & 11.033 & 0.933 & 11.500 \\
PARWiS & \textbf{0.833} & 7.760 & \textbf{0.833} & 7.760 & \textbf{0.833} & 7.760 \\
Contextual PARWiS & \textbf{0.833} & 7.760 & \textbf{0.833} & 7.760 & \textbf{0.833} & 7.760 \\
RL PARWiS & 0.900 & \textbf{7.296} & 0.900 & \textbf{7.296} & 0.900 & \textbf{7.296} \\
\bottomrule
\end{tabular}
\end{table*}

\begin{table*}[h]
\centering
\caption{Selected t-test results (Double TS vs. PARWiS, Recovery) across datasets and budgets.}
\label{tab:ttest_recovery_comparison}
\begin{tabular}{lccccccccc}
\toprule
& \multicolumn{3}{c}{Synthetic} & \multicolumn{3}{c}{Jester} & \multicolumn{3}{c}{MovieLens} \\
\cmidrule(lr){2-4} \cmidrule(lr){5-7} \cmidrule(lr){8-10}
Budget & \( B=40 \) & \( B=60 \) & \( B=80 \) & \( B=40 \) & \( B=60 \) & \( B=80 \) & \( B=40 \) & \( B=60 \) & \( B=80 \) \\
\midrule
t-stat & -2.246 & -3.862 & -1.616 & -2.594 & -1.921 & 0.000 & -0.356 & -1.201 & -1.201 \\
p-value & 0.029 & 0.000 & 0.112 & 0.012 & 0.060 & 1.000 & 0.723 & 0.235 & 0.235 \\
\bottomrule
\end{tabular}
\end{table*}

\subsection{Code, Python Package, and Data Availability}
The source code for the Dueling Bandit Toolkit, including implementations of the PARWiS, Contextual PARWiS, RL PARWiS, Double Thompson Sampling, and Random Pair baseline algorithms, is publicly available on GitHub at \url{https://github.com/shailendrabhandari/dueling_bandit}. The repository contains the full codebase, documentation, and example scripts to reproduce the experiments presented in this paper. The toolkit is also distributed as a Python package on the Python Package Index (PyPI: \url{https://pypi.org/project/dueling-bandit/}) and can be installed using pip:

\begin{verbatim}
pip install dueling-bandit
\end{verbatim}

Comprehensive documentation, including API references, tutorials, and experimental details, is hosted on ReadTheDocs at \url{https://dueling-bandit.readthedocs.io/en/latest/}. The documentation provides step-by-step guides for setting up the toolkit, running simulations, and visualizing results, as well as a detailed description of the algorithms and datasets used in this study.

\end{document}